\documentclass[twocolumn]{article}

\title{How to See Hidden Patterns in Metamaterials with Interpretable Machine Learning}
\usepackage{authblk}
\usepackage[symbol]{footmisc}
\author[1]{Zhi Chen\footnote{Corresponding author. E-mail: zhi.chen1@duke.edu}}
\author[2]{Alexander Ogren}
\author[2]{Chiara Daraio}
\author[1]{L. Catherine Brinson}
\author[1]{Cynthia Rudin}
\affil[1]{Duke University, NC 27708, USA}
\affil[2]{California Institute of Technology, CA 91125, USA}
\date{}

\usepackage[utf8]{inputenc} 
\usepackage[T1]{fontenc}    
\usepackage{url}            
\usepackage{booktabs}       
\usepackage{amsfonts}       
\usepackage{nicefrac}       
\usepackage{microtype}      
\usepackage{graphicx}
\usepackage{subcaption}
\usepackage[section]{algorithm}
\usepackage{algpseudocode}
\usepackage{amsmath}
\usepackage{amssymb}
\def\tree{\textrm{tree}}
\usepackage{bbold}
\usepackage[numbers]{natbib}
\usepackage{wrapfig}
\usepackage[english]{babel}
\usepackage[utf8]{inputenc}
\usepackage{amsthm}
\usepackage{optidef}
\usepackage{makecell}
\usepackage{multirow}
\usepackage{float}
\usepackage{bm}
\usepackage{xcolor}
\usepackage{hyperref}       

\providecommand{\keywords}[1]
{
  \small	
  \textbf{\textit{Keywords---}} #1
}

\def\PrototypeModel{unit-cell template set}
\def\Prototype{unit-cell template}

\begin{document}

\maketitle
\begin{abstract}

Machine learning models can assist with metamaterials design by approximating computationally expensive simulators or solving inverse design problems. However, past work has usually relied on black box deep neural networks, whose reasoning processes are opaque and require enormous datasets that are expensive to obtain. In this work, we develop two novel machine learning approaches to metamaterials discovery that have neither of these disadvantages. These approaches, called \textit{shape-frequency features} and \textit{unit-cell templates}, can discover 2D metamaterials with user-specified frequency band gaps. Our approaches provide logical rule-based conditions on metamaterial unit-cells that allow for interpretable reasoning processes, and generalize well across design spaces of different resolutions. The templates also provide design flexibility where users can almost freely design the fine resolution features of a unit-cell without affecting the user's desired band gap. 
\end{abstract}

\keywords{Phononic materials, Frequency Band Gaps, Interpretable machine learning, Rule-based models}

\section{Introduction}

\begin{figure*}[th]
    \centering
    \includegraphics[width=6in]{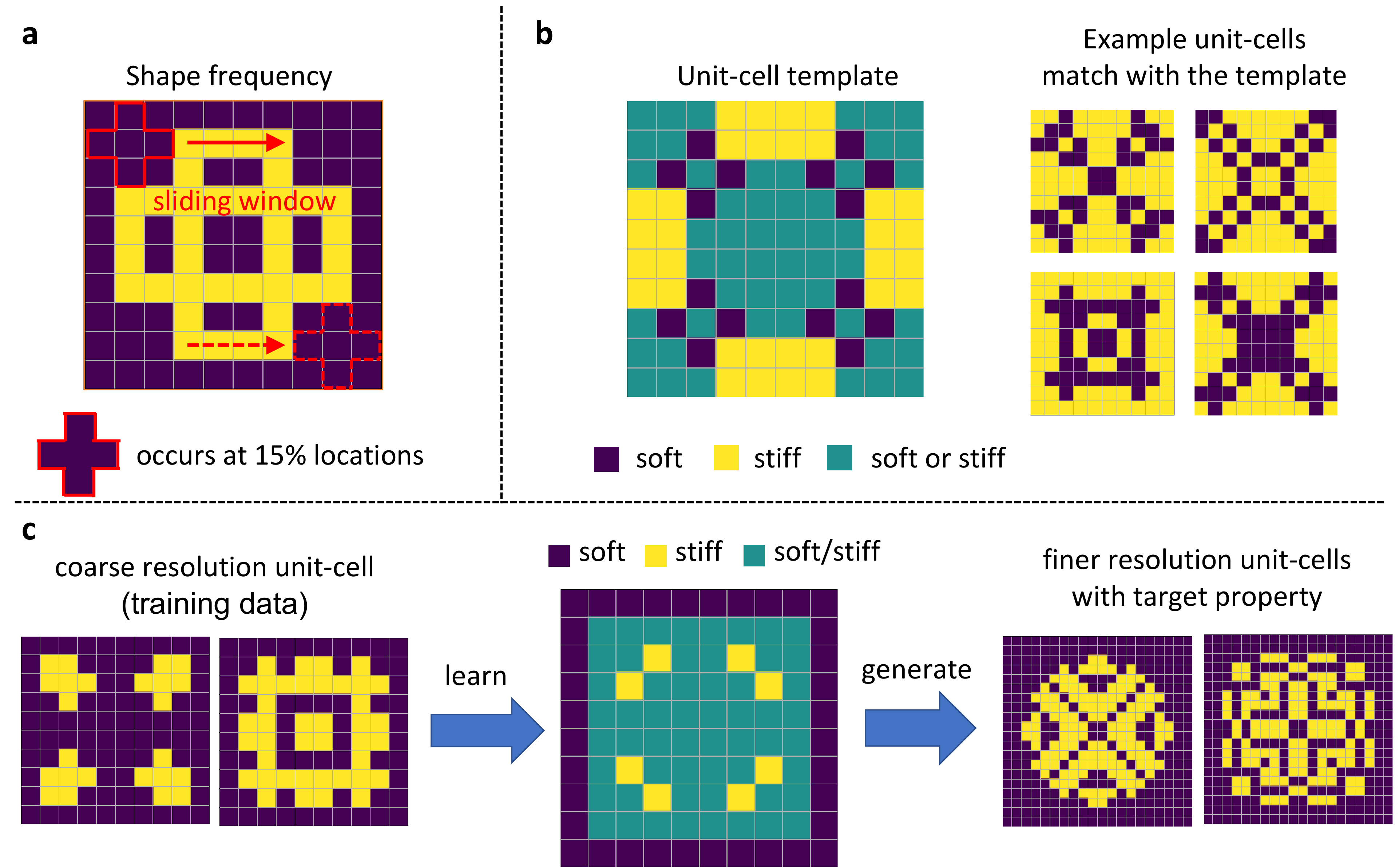}
    \caption{Examples of interpretable key patterns discovered by the proposed method. \textbf{a}. A shape frequency feature (this one is shaped like a ``+'').   How frequently this shape appears in the unit-cell is a useful predictor of a band gap. \textbf{b}. A unit-cell template, which considers specific global patterns in the unit-cell. 
    Here, regardless of whether we place stiff or soft materials at each green pixel in the unit-cell, as long as the stiff and soft materials are in the positions defined by the template in yellow and purple, there will be a band gap within the user's desired range. \textbf{c}. The patterns are learned from coarse resolution training data but can be robustly transfered to finer resolution, and generate fine resolution unit-cells with the target property.}
    \label{fig:intro}
\end{figure*}
Metamaterials are traditionally designed through empirical trial-and-error or intuition \citep{38} or computationally expensive topology optimization \citep{sigmund2003systematic,37,Bilal_PRE_2011,38,39} which often produces designs out of predetermined geometrical building blocks. Recently, machine learning methods (ML) have gain popularity for metamaterial design. For example, machine learning models can be trained to predict material properties from unit cells defined by a finite set of pixels/voxels, and used as faster surrogate models for computationally expensive simulations \cite{nadell2019deep,bostanabad2019globally,qiu2019deep,finol2019deep,wiecha2019deep}. Using deep generative models and neural network inversion techniques, some recent works \cite{liu2018generative,ma2018deep,deng2021neural,ma2019probabilistic,wang2020deep,ren2020three,liu2020compounding} aim to directly solve the inverse design problem for metamaterials, i.e., generating the designs given the target property. However, the ML models used in these work are usually gigantic black boxes, whose decision processes are hard to understand. This is undesirable for scientific discovery purposes because scientists may also want to gain insights into what geometric features are important for a given target property, such a particular frequency band gap. In addition, these massive models are data hungry and not robust to distributional shift -- they usually require a huge simulated dataset that covers most of the design space. These models also tend to perform poorly on datasets they have not seen before, such as unit-cells in a finer resolution space.

In this paper, instead of relying on existing black box approaches, we propose two novel rule-based ML approaches for metamaterial design that have major advantages:
\begin{itemize}
\item \textbf{\textit{Interpretability:}} The approach allows us to discover interpretable key patterns within unit-cells that are related to a physical property of interest (see Figure \ref{fig:intro}\textbf{a} and \ref{fig:intro}\textbf{b}). We consider two types of patterns: (i) local patterns called \textit{shape frequency features}, which calculate the occurrence frequency of certain shapes in the unit-cell; (ii) global patterns, called \textit{unit-cell templates}, which look for arrangements of constituent materials in specific regions of the metamaterials' unit-cells. The unit-cell templates are optimized with binary integer programming to find global patterns within unit cells that give the metamaterial a desired property.
\item \textbf{\textit{Leverages Multi-resolution Properties:}} An important observation underpinning our methodology is that a pattern in the coarser resolution design space also exists in finer resolution design space, with one coarse pixel replaced by many finer pixels. As a result, \textit{if a pattern can robustly characterize the target property at the coarse resolution design space, it will also be predictive at the finer resolution design space}. This leads to computationally-efficient discovery of many valuable metamaterial designs possessing the desired properties. In particular, our method allows us to construct a \textit{scaffold of patterns} 
that allows interpretable coarse scale information discovered at low resolutions to be reliably transferred to make accurate predictions for high resolution designs (see Figure \ref{fig:intro}\textbf{c}). 

\item \textbf{\textit{Flexible Metamaterial Designs:}} Our unit-cell templates (e.g., the one in Figure \ref{fig:intro}\textbf{b}) enables flexibility in unit-cell designs at any resolution. Simply, unit-cell templates specify regions where one can almost freely design unit-cell features without changing the target band gap property (e.g., in Fig \ref{fig:intro}\textbf{b} constituent phases can be arranged at will in the green regions determined by our algorithm). Such flexibility in design might be useful to satisfy practicality constraints such as connectivity or other design constraints such as overall stiffness.

\end{itemize}

Section \ref{sec:key_problems} discusses related works on ML approaches for metamaterial designs. In Section \ref{sec:settings}, we introduce the problem setting. In Section \ref{sec:method}, we provide the proposed methods and explain how they deal with the four core objectives of data-driven approaches
that are important to materials scientists: (a)
design-to-property prediction; (b) property-todesign
sampling; (c) identify key patterns; (d)
transfer to finer resolution. We then evaluate performance of the proposed methods in Section \ref{sec:experiments} and test them on practical applications. We discuss and conclude in Section \ref{sec:conclusion}.

\section{Related Works}
\label{sec:key_problems}



Metamaterials are architected materials with engineered geometrical micro- and meso-structures that can lead to uncommon physical properties.
As mentioned in the introduction, many existing works apply machine learning models for designing metamaterials, i.e., assembling constituent materials into metamaterials that have specific physical properties. 

Much past work focuses on structure-to-property prediction. Because of the expensive computational cost of numerical simulations, these works train machine learning models (mostly deep learning models) to approximate the simulation results \cite{nadell2019deep,bostanabad2019globally,qiu2019deep,finol2019deep,wiecha2019deep}. Using these ML models as a fast replacement of the simulator,  materials satisfying the design objective can be found more efficiently using rejection sampling, i.e., randomly picking a structure in the design space until it satisfies the design objective. However, 
given the immense size of the design space, finding materials through rejection sampling can still be computationally inefficient. Therefore, some recent works apply deep generative models and neural network inverse modeling \cite{liu2018generative,ma2018deep,deng2021neural,ma2019probabilistic,wang2020deep,ren2020three,liu2020compounding} to train a more efficient materials sampler, aiming to solve the inverse design problems for metamaterials, i.e., property-to-structure sampling. See \cite{jiang2020deep,ma2021deep,xu2021interfacing} for reviews on using deep learning methods for metamaterial designs.

Almost all modern existing work on this topic uses black box models like deep neural networks. Such approaches are unable to answer key questions such as ``What patterns in a material's design would lead to a specific desirable property?'' A link between the specific design and the target property could be useful for further research; i.e., to determine whether there is an agreement with domain knowledge, and if not, to potentially discover new knowledge.

Other work also stresses the importance of interpretability  for metamaterial design \cite{ma2019probabilistic,elzouka2020interpretable,zhu2022harnessing}. These works are very different from ours, and cannot solve the challenge we want to address. Ma et al$.$ \cite{ma2019probabilistic} try to make the latent space of deep generative models interpretable. Their interpretable features describe general geometrical information such as size and shape of holes in the material but these properties are not associated with the target property. In contrast, our goal is to find key patterns within unit cells that result in the target property (in our case, a band gap). Elzouka et al$.$ \cite{elzouka2020interpretable} and Zhu et al$.$ \cite{zhu2022harnessing} also build rule-based models, namely decision trees, but the design problems they are solving are different and much simpler than the problem we try to solve. Specifically, their materials are described by several continuous features, e.g., thickness of the materials, while we work on pixelated metamaterials whose features are just raw pixels made of constituent materials. Discovering interpretable patterns directly from raw pixels is a more challenging and fundamental problem.

Another serious issue with using black box models like deep neural networks is that they require large labeled training datasets, and the training and testing data should come from the same distribution so that the model generalizes between training and test. However, constructing these large datasets is extremely expensive because the labels (i.e., material properties) of metamaterials are calculated by simulation, which is computationally expensive. In fact, this whole process could be so expensive that one might find it less expensive to use the simulator to get the results on the test set directly rather than go through the process of collecting a training set at all.
Note that deep generative models like GANs \cite{goodfellow2014generative, creswell2018generative} do not address the simulation bottleneck because training a GAN requires co-training a generator and a discriminator which requires even more training data than just training a structure-to-property predictor. 
Ma et al$.$ \cite{ma2020data} propose a self-supervised learning approach that can utilize randomly generated unlabeled data during training to reduce the amount of training data. However, this work can only handle structure-to-property prediction but not the problem of inverse design considered here. Our method instead alleviates this simulation bottleneck through a \textit{multi-resolution} approach: our unit-cell template models are trained using a (relatively small amount of) \textit{coarse-resolution} data but can extrapolate and generate a (large amount of) \textit{finer-resolution} metamaterial designs. This is helpful because the coarse resolution space is much smaller than the fine-resolution space, and gives us a bird's eye view of what might happen when we sample at the finer scale throughout the space of possible metamaterials.

\section{Problem Settings} 
\label{sec:settings}

Here, we introduce the settings of the metamaterial design problem we are trying to solve, including the inputs and target of the dataset and their physical meanings.

\begin{figure*}
    \centering
    \includegraphics[width=6.2in]{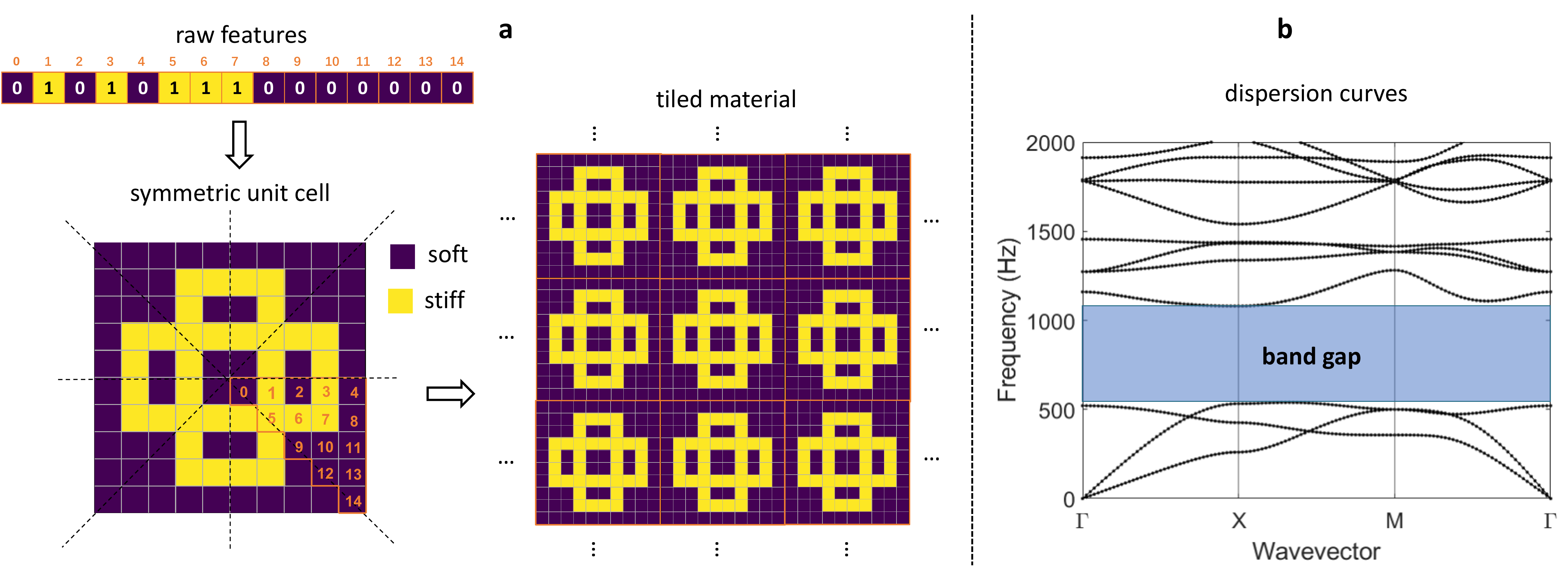}
    \caption{Input and target of our metamaterial design problem. \textbf{a.} 2-D phononic material constructed with the raw features. Upper Left: raw 15 dimensional input feature vector. Lower left: the feature vector defines the triangle in the lower right of the unit-cell. The triangle is copied using lines of symmetry to define the full unit-cell. Right: the unit-cell is tiled to obtain the full material. For the Bloch-Floquet boundary conditions, the tiling is infinite in all dimensions; \textbf{b.} The design objective is the frequency band gaps.}
    \label{fig:unitcell}
\end{figure*}

We aim to design and characterize 2-D pixelated metamaterials made by tiling a $10\times 10$ unit-cell. Such materials can be stacked to form 3D structures, and can direct, reflect or scatter waves, depending on the choice of unit-cell's material selection and geometry. In our framework, the unit-cell is a square with side length $a$ = 0.1 m. 
However, transferring to a different length scale for a different application is easily doable by a simple scaling transformation on the dispersion relations. For a $10\times 10$ unit-cell, the pixel side length is 1 cm; for $20\times 20$ unit-cell, the pixel side length is 0.5 cm. Each unit-cell is made of two constituent materials: one is soft and lightweight, with elastic modulus $E$ = 2 GPa\footnote{GPa is gigapascals, a unit used to quantify elastic modulus.} and density $\rho$ = 1,000 kg/m$^3$, and the other is stiff and heavy with $E$ = 200 GPa, and $\rho$ = 8,000 kg/$m^3$. These two sets of material properties are representative of a polymer and steel respectively.
Our unit-cells are symmetric, with four axes of symmetry ($x$,$y$ and $\pm 45^{\circ}$). Under the symmetry constraints, the coarsest resolution ($10\times 10$) unit-cell has only 15 irreducible pixels. As a result, the raw input features of a sample in our dataset is a 15-dimensional binary vector: 0 means the soft constituent material in that location, and 1 means stiff constituent material. Thus, the full coarse space can be characterized, having $2^{15}$ total states. Figure \ref{fig:unitcell} shows how to construct a material from the representation involving the 15 raw input features.

The material property we desire in our engineered materials is a band gap within a specific frequency range, given by the user. A band gap is a range of frequencies within which elastic waves cannot propagate and are instead reflected. 

To identify the existence of a band gap, one can examine the effect of dispersion in metamaterials by calculating dispersion relations. Dispersion relations are functions that relate the wavenumber of a wave to its frequency, and they contain information regarding the frequency dependent propagation and attenuation of waves. Dispersion relations are found by computing elastic wave propagation solutions over a dense grid of wavevectors. A band gap exists when there is a range of frequencies in the dispersion relation for which no wave propagation solutions exist (see Figure \ref{fig:unitcell} \textbf{b}).

Dispersion relation computations use Bloch-Floquet periodic boundary conditions, i.e., they assume that a given unit-cell is tiled infinitely in space. The physics revealed in dispersion analysis (infinite-tiling) can be leveraged in more realistic finite-tiling scenarios, as we will demonstrate later, which makes dispersion relation computations very useful for exploration and design of real materials. 
Our dispersion relation simulations are implemented using the finite element method. 

 
More details about the simulation can be found in the Supplementary Information A. 

We are looking for materials with band gaps in a certain frequency range. To define this task as a supervised classification problem, we create a binary label based on existence of a band gap in a given frequency range (e.g. [10, 20] kHz): 1 means one or more band gaps exist, 0 means no band gap exists. In other words, if a band gap range intersects with the target frequency range, the label is 1, otherwise the label is 0. One can also flexibly adjust the band gap label for different practical uses. For example, we can set the label to 1 only when the intersection of the band gap and the target frequency range is above a minimum threshold. We can also set the label to 1 when the band gap covers the entire target range, or even create a label for band gap properties in multiple frequency ranges.


\section{Method}
\label{sec:method}
This section introduces proposed methods. Section \ref{sec:sff} explains the shape-frequency features and how they can be used to optimize different objectives. Inspired by the efficiency of shape-frequency features, we then propose unit-cell template sets in Section \ref{sec:prototype}. 

\subsection{Shape-Frequency Features}
\label{sec:sff}
We hypothesize that the occurrence of certain local features in the metamaterials might contribute to the formation of band gaps. For example, certain local patterns/shapes in the materials can lead to interference and thus cause band gaps. Because the unit-cells are repeated, the location of such local patterns does not matter, \textit{as long as they occur frequently}, the band gap can be formed. Such physics intuition inspires us to propose \textit{shape-frequency features} which calculates the \textit{number of times a pattern occurs in the unit cell divided by total number of locations}. 

Denote the unit-cell as a $n\times n$ binary matrix $\mathbf{U}\in \{0,1\}^{n\times n}$, where $\mathbf{U}_{i,j}=0$ means pixel $i,j$ is assigned to the soft material and $\mathbf{U}_{i,j}=1$ when the pixel is assigned to the stiff material.

A specific shape $s$ can be represented as a set of location offsets $O_s$ whose elements are coordinates of pixels with respect to a reference pixel. 
For example, a $2\times 2$ square window can be represented as $\{(0,0),(1,0),(0,1),(1,1)\}$. A $3\times 3$ plus symbol as in Figure \ref{fig:intro} can be represented as $\{(0,1), (1,0), (1,1), (1,2), (2,1)\}$. For a specific unit-cell, the feature value corresponding to that shape is computed by sliding the shape
over the unit-cell and calculating the fraction of times that it is entirely contained within the soft material,
\begin{equation}
    f_s = \frac{1}{n^2}\sum_{i=1}^{n}\sum_{j=1}^{n} \mathbb{1}\left[\left(\sum_{(o_r,o_c)\in O_s}\mathbf{U}_{i+o_r,j+o_c}\right)=0\right],
\end{equation}
where $\mathbb{1}[\cdot]$ is the indicator function. It equals 1 if and only if all the pixels in the shape are soft material (i.e., $\sum_{o_r,o_c}\mathbf{U}_{i+o_r,j+o_c}=0$).

We would typically consider a collection of shapes, and have one element in a unit-cell's feature vector per shape. Thus, for unit-cell $i$, the $j$th component of its feature vector corresponds to how often the full shape $j$ appears in its soft material. 
\begin{figure*}
    \centering
    \includegraphics[width=5.in]{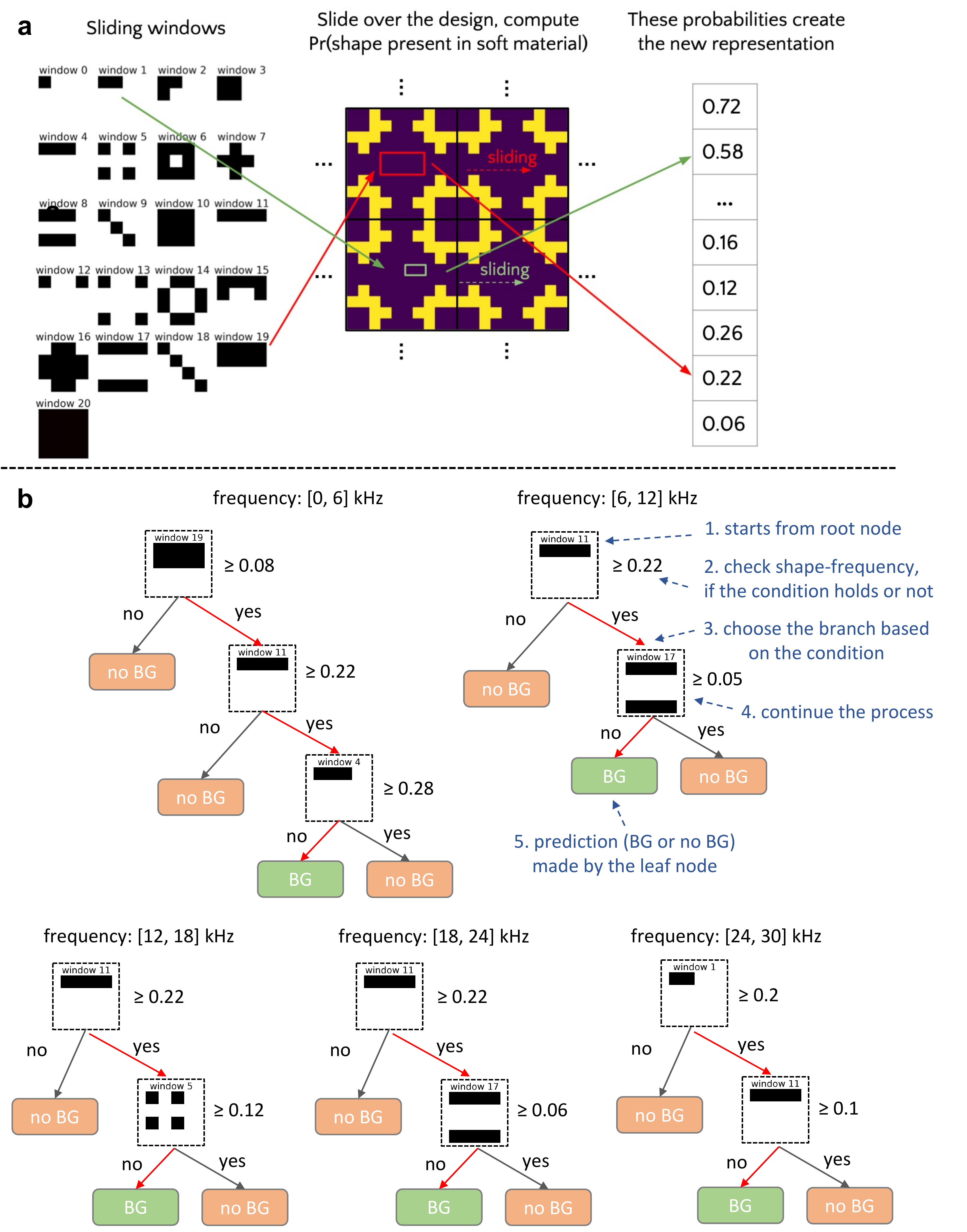}    \caption{\textbf{a.} The process of calculating shape-frequency features. \textit{Left}: Collection of shapes (sliding windows) used to create the shape-frequency features. \textit{Middle:} An example $10\times 10$ unit-cell design (tiled 4 times for better visualization). \textit{Right:} Shape-frequency features of the unit-cell, which count the fraction of locations in the unit-cell where the shape is present in the soft material; \textbf{b.} Optimal sparse decision trees built on shape-frequency features for predicting band gaps in different frequency ranges; red arrows denote the paths to band gap (BG) nodes. Blue text on the top-right tree breaks down how the decision tree predicts whether the band gap exists using shape-frequency features.}
    \label{fig:shapes}
\end{figure*}

In more detail, consider the collection of shapes, shown within Figure \ref{fig:shapes}\textbf{a} (left). Note that this collection of shapes is the full set used in the results section, not just examples of them. Again, pixels of the stiff material are in yellow, and the soft material is shown in purple. We slide each of the shapes (sliding windows) over the unit-cell (Figure \ref{fig:shapes}\textbf{a} (middle))
and count the fraction of positions over which the shape is fully contained within pixels of the soft (purple) material. These fractions together form the new representation for the unit-cell (Figure \ref{fig:shapes}\textbf{a} (right)). Note that, to calculate the fraction, we should also consider the situation where the sliding window is across the boundary of two unit-cells, since the entire material is made by tiling the unit-cell.
Because the unit-cells are symmetric, the occurrence of the patterns within the unit-cell are also symmetric: if we rotated the patterns by 90$^{\circ}$, 180$^{\circ}$ or 270$^{\circ}$, the number of detections of the pattern within the unit-cell would be identical. Note that the shape-frequency features are different from standard convolution filters used in computer vision; details are discussed in the Supplementary Information B. Theoretically, our method can be generalized to nonsquare pixels and unit-cells as well, see Supplementary Information E.

Once we calculate the shape-frequency features of the unit-cells, they can replace the original raw features and be used as the inputs of the machine learning models to predict the band gap output. We show later in Section \ref{sec:experiments:prediction} that using the shape-frequency features as inputs, machine learning models can predict the existence of band gaps more accurately than using raw features. 

Since shape-frequency features are just new representations of the unit-cells and they are written in vector form, any type of machine learning model can be trained to predict band gap existence, taking these features as inputs. These machine learning models can not only be complex models like neural networks or boosted trees, but can also be interpretable models (e.g., sparse decision trees).  Figure \ref{fig:shapes}\textbf{b} shows examples of sparse decision trees that predict the existence of band gaps from shape frequency features. When making predictions, the decision tree starts from its root node, checking if the shape shown on the node appears frequently in the unit-cell (e.g., if a 1$\times$4 soft bar occurs in the unit-cell more than 22\% of all possible locations). If so, it goes to the right branch; if not it goes to the left branch. The process is continued until a leaf node (in green or orange) is reached. At that point, it outputs the prediction of whether a band gap exists, based on the majority vote of the training data within that leaf. The paths denoted by red arrows in the trees in Figure \ref{fig:shapes}\textbf{b} show how often the local patterns need to occur in the unit-cells to predict an open band gap. More analysis of these discovered patterns can be found in Section \ref{sec:experiments:key_patterns}.

\subsubsection{Optimizing Precision and Support}
\label{sec:gosdt_objective}
For regular binary classification problems, we hope the machine learning models have high accuracy for both positive samples (materials with band gaps) and negative samples (materials with no band gap). This is also the objective if our goal is to perform only structure-to-property prediction. However, for the property-to-structure task, where our goal is to produce a number of unit-cell designs with the target band gap property, prediction accuracy is no longer a good metric. Instead, we hope all unit-cells that are predicted to have a band gap actually have a band gap, i.e., we prefer that the model has high \textit{precision}. We also hope the total number of discovered designs, i.e., the \textit{support}, meets the requirement of real applications. Thus, for the property-to-structure task, our objective is a combination of precision and support.

Most machine learning methods cannot directly optimize custom objectives with constraints, such as precision, constrained by support. However, there are new approaches that permit direct optimization of custom discrete objectives. We use GOSDT (Generalized and Scalable Optimal Sparse Decision Trees, \cite{GOSDT}) for this task, because it directly optimizes decision trees for customized objectives. Different from traditional decision tree algorithms which use greedy splitting and pruning, GOSDT directly searches through the space of all possible tree structures, uses analytical bounds to reduce a huge amount of search space, and directly outputs a tree that optimizes the customized objective. We programmed it to maximize the following custom objective to optimality:
\begin{eqnarray}
\lefteqn{\max_{\tree} \left[
\frac{TP_{\tree}}{TP_{\tree}+FP_{\tree}+\epsilon} -  \frac{K}{TP_{\tree}+\epsilon}\right]}
\label{eqn:gosdt_objective_1} \\
&= \max_{\tree} \left[
\frac{P-FN_{\tree}}{P-FN_{\tree}+FP_{\tree}+\epsilon} -  \frac{K}{P-FN_{\tree}+\epsilon}\right]. \label{eqn:gosdt_objective_2}
\end{eqnarray}
Here, $K$ is a parameter that balances the precision and the support; $\epsilon$ is a small constant for numerical convenience; $TP$, $FP$, $TN$, $FN$ mean true positives, false postives, true negatives and false negatives. Equation (\ref{eqn:gosdt_objective_1}) shows precision (first term) and inverse support (second term); we use inverse support so that if support is large, the term diminishes in importance. The simplification in Equation (\ref{eqn:gosdt_objective_2}) shows that the objective is monotonically decreasing with respect to $FN_{\tree}$ and $FP_{\tree}$. Theorem B.1 of \citep{GOSDT} shows that as long as the objective is decreasing with respect to $FN_{\tree}$ and $FP_{\tree}$, we can find an optimal sparse decision tree using GOSDT's branch and bound algorithm. 

\subsubsection{Property-to-structure Sampling}
\label{sec:sff:p2d}
 Because we optimize the precision using GOSDT, the false positive rate of our model will be low enough to work with. At this point, we directly do rejection sampling using the decision tree to produce unit-cell designs with the target band gap. Specifically, to produce valid designs, the rejection sampling approach randomly picks structures in the design space, evaluates whether each of them are predicted to have a band gap, and outputs only these relevant designs.
 After sampling, each accepted sample is evaluated with the physics-based finite-element model to determine whether a band gap is present.
 
\subsubsection{Transfer to Finer Resolution}
\label{sec:sff:transfer}

\begin{algorithm*}[ht]
 \caption{Sampling Fine resolution Unit-cell Designs via Shape-frequency Features}
 \begin{tabbing}
 xxx \= xx \= xx \= xx \= xx \= xx \kill
 \textbf{Input}: simulated coarse resolution ($10\times 10$) dataset $\mathcal{D}:=\{\mathbf{x}_i,y_i\}_{i=1}^{2^{15}}$, $\mathbf{x}_i\in \{0,1\}^{15}$: raw features; $y_i$: band gap label\\
 \textbf{Parameters:} set of shapes $S$; tree sparsity regularization $\lambda$ (see \cite{GOSDT}); $K$, $\epsilon$ (see Section \ref{sec:gosdt_objective}) \\
 \textbf{Output:} raw features of a fine resolution unit-cell $\mathbf{\tilde{x}}$\\
 1: \> calculate shape-frequency features $\mathbf{x}_i^{\text{SFF}}$ = SFF($\mathbf{x}_i$,$S$) for all $\mathbf{x}_i$ in the dataset (coarse resolution), see Section \ref{sec:sff}\\
 2: \> train an optimal sparse decision tree $\tau$ = GOSDT($\{\mathbf{x}_i^{\text{SFF}},y_i\}_{i=1}^{2^{15}}$, $\lambda$, $K$, $\epsilon$), see Section \ref{sec:sff:p2d} \\
 3: \> \textbf{while}(True): \\
 4: \> \> randomly sample a binary vector $\mathbf{\tilde{x}}$ as raw features in the fine resolution space \\
 5: \> \> calculate shape-frequency features $\mathbf{\tilde{x}}^{\text{SFF}}$ = SFF($\mathbf{\tilde{x}}$,$S$) (fine resolution), see Section \ref{sec:sff:transfer} \\
 6: \> \> \textbf{if} $\tau(\mathbf{\tilde{x}}^{\text{SFF}})=1$: \\
 7: \> \> \> \textbf{return} $\mathbf{\tilde{x}}$
 \end{tabbing}
 \label{alg:sff_gosdt}
\end{algorithm*}
The raw feature space of finer resolution samples is different from that of coarse resolution samples. Therefore, for the finer resolution data, we need to slightly modify the approach to obtain shape-frequency features that are compatible with the coarse resolution shape-frequency features. Suppose we want to transfer the model from $10\times 10$ to $20\times 20$ space. Then, there are three changes in calculating the shape-frequency features:
\begin{itemize}
    \item The window size should be doubled when moving from coarse resolution to fine resolution;
    \item The stride of the sliding window should be 2 instead of 1;
    \item In counting the shape-frequency values, exact agreement between the window and soft material (purple) should be replaced with near exact agreement. In particular, when less than 2 yellow pixels (stiff material) are found in the window, we can consider this to be an agreement.
\end{itemize}

When using the shape-frequency features with these modifications, the decision tree model learned on the coarse resolution data can be directly applied to the fine resolution. Thus, we can also do rejection sampling on the fine resolution with the model.

Algorithm \ref{alg:sff_gosdt} shows the entire pipeline of using an optimal sparse decision tree built on shape-frequency features to sample fine resolution designs with the target band gap property. Visual illustration of the sampling process can be found in Supplementary Information C.

\subsection{Unit-cell Template Sets}
\label{sec:prototype}
\begin{figure*}
    \centering
    \includegraphics[width=5.5in]{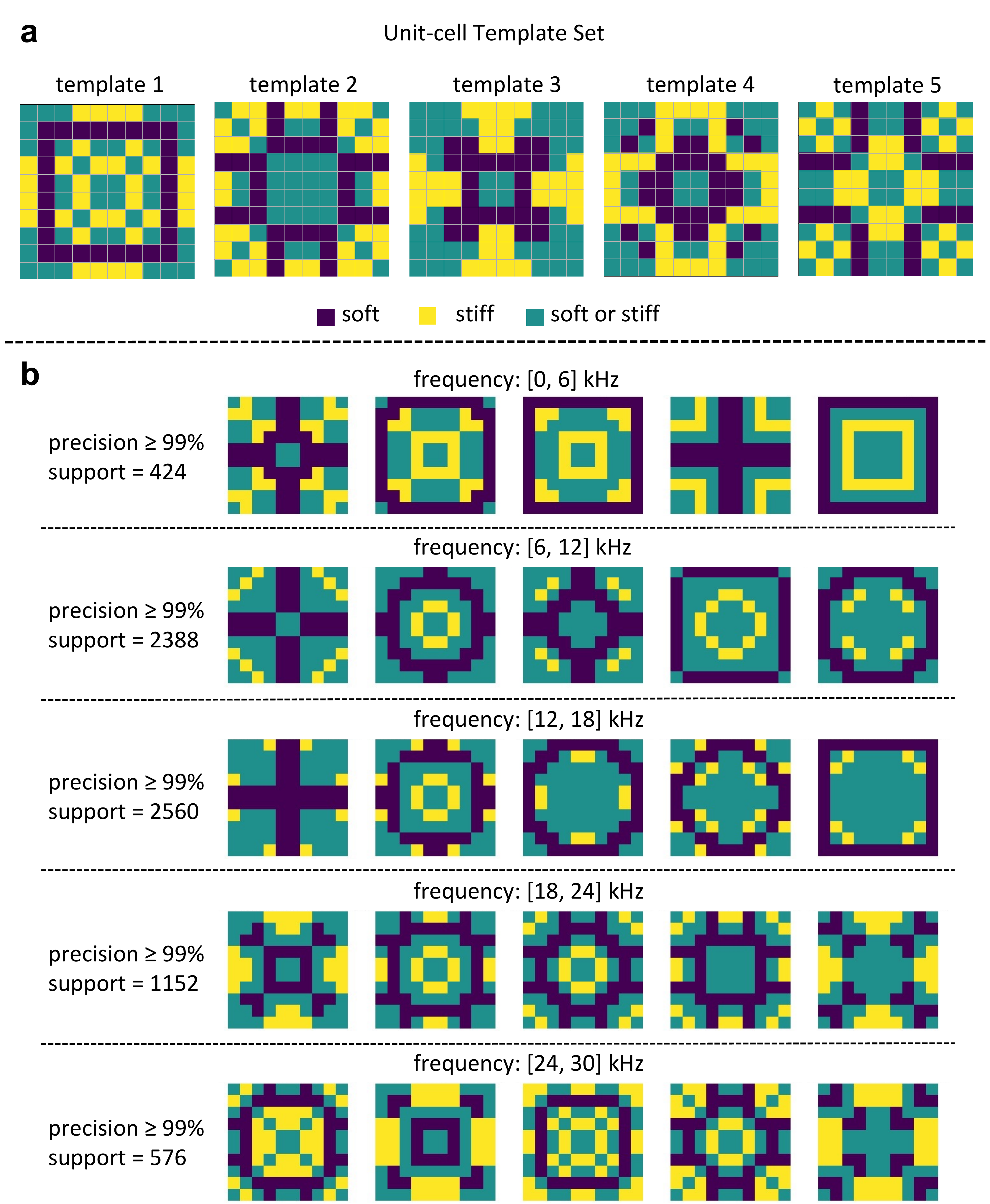}
    \caption{\textbf{a}. An Example of a \PrototypeModel{}. A unit-cell is predicted as positive as long as it matches at least one \Prototype{} in the set; \textbf{b}. \PrototypeModel{s} learned for predicting band gaps in different frequency ranges. Note that because unit-cells are tiled, a large cross through the center is identical to a square on the border. (E.g., consider the two upper right unit-cells.)}
    \label{fig:unitcell_template_set}
\end{figure*}

Here, we introduce another interpretable machine learning model, called \PrototypeModel{s}. Different from sparse trees on shape-frequency features, which focus on local patterns, a \Prototype{} captures a global pattern for the unit-cell that is related to the target properties. The \Prototype{} is a $n\times n$ matrix $\mathbf{T}\in \{0,1,*\}^{n\times n}$, where $\mathbf{T}_{i,j} = 0$ means the pixel is soft material, $\mathbf{T}_{i,j} = 1$ means the pixel is stiff material, and $\mathbf{T}_{i,j} = *$ means the pixel could be either soft or stiff, i.e., a free pixel. 

\noindent \textbf{Definition} \textit{(match)}. We say a unit-cell design $\mathbf{U}$ \textit{matches} the \Prototype{} if and only if all pixels with value 0 on the template are also 0 on the design, and all pixels with value 1 on the \Prototype{} are also 1 on the design. That is, $\forall (i,j)$ such that $\mathbf{T}_{i,j} \neq *$, we have $\mathbf{U}_{i,j} = \mathbf{T}_{i,j}$.

A \PrototypeModel{} contains a set of \Prototype{s}, and the sample design is predicted as positive if and only if it matches at least one \Prototype{} in the set. Figure \ref{fig:unitcell_template_set}\textbf{a} shows an example of a \PrototypeModel{} that consists of five different templates. We proposed \Prototype{s} because we found that some pixels in the unit-cells are more important for the formation of band gaps than others. For the pixels that are not important, even if they are flipped, the band gaps remain unchanged; we denote these as free pixels in the \Prototype{}. The free pixels identify unimportant regions where changes do not affect the target band gaps, while the other pixels form the key global pattern that leads to the band gap. We aim to find a \PrototypeModel{} that captures a diverse set of global patterns related to the target band gap. The relationships between \PrototypeModel{s} and other machine learning methods are discussed in  Supplementary Information B. Theoretically, our method can be generalized to nonsquare pixels and unit-cells as well, see Supplementary Information E.

The training objective of the model is to find a small number of \Prototype{s}, such that the training precision of the entire model is high enough, and the model covers as many valid designs as possible, i.e., maximizing the support under a minimum precision constraint. The reasons for optimizing precision and support were explained in Section \ref{sec:gosdt_objective}, and we set a limit for the total number of selected \Prototype{s} to encourage the \PrototypeModel{s} to contain a diverse set of \Prototype{s}. Because the total number of possible templates is extremely large, the training process is divided into two steps, a pre-selection of candidate \Prototype{s} that filters out useless templates to reduce the problem size (Section \ref{sec:preselect}), and a integer linear programming (ILP) formulation to optimally select from the candidates obtained in the first step (Section \ref{sec:ilp}). Figure \ref{fig:unitcell_template_set}\textbf{b} shows the \PrototypeModel{s} learned by the proposed algorithm for band gap prediction. More analysis of these discovered patterns can be found in Section \ref{sec:experiments:key_patterns}.

\subsubsection{Pre-selection of Templates}
\label{sec:preselect}
Here we consider symmetric \Prototype{s}, because designs in our dataset are all symmetric. By removing all symmetry redundancy, the \Prototype{} can be represented by its irreducible pixels, i.e. a 15 dimensional vector $\mathbf{t}\in \{0,1,*\}^{15}$. The total number of possible templates is $3^{15}$ which is approximately 14.3 million. This is too large for the ILP in the next step. However, among the $3^{15}$ possible templates, most of them would never be selected because either their precision or support is not high enough. For example, a \Prototype{} with precision 80\% is not likely to be used if we want the entire model to have precision above 99\%, i.e., we hope the unit-cells are all having the desired band gap properties. Also, if the support of a \Prototype{}, i.e., the total number of designs that match it, is very small (e.g., $<$ 10), the model may not generalize well. Therefore, we pre-select the \Prototype{s} by setting minimum thresholds of precision and support, which reduces the search space only to promising \Prototype{s}. We select the \Prototype{} as a candidate only when it meets the minimum thresholds.

Observing that the entries are all binary in the unit-cell designs, we implement the precision calculation via bit operations, which significantly improves the speed of the pre-selection step. With the bit-operation implementation, the pre-selection steps of all $3^{15}$ possible templates finish in 50 seconds. The number of \Prototype{s} that remains is typically in the range of 6000 to 12000, which is now suitable for ILP.

\subsubsection{ILP for Template Selection}
\label{sec:ilp}
After the pre-selection step, we have a set of candidate \Prototype{s}. Since the pre-selection step significantly cuts down the space of templates, we can directly formulate the template selection as an optimization problem, and solve it to provable optimality. Specifically, we formulate a ILP to optimally select from the candidates. Suppose we have $n$ designs and $m$ candidate templates. The goal of the ILP is to choose at most $s$ \Prototype{s} ($s\ll m$) whose union forms a model, such that the support is maximized and the precision of the model is at least $p$. Denote the true labels of all designs by a binary vector $\mathbf{y}\in \{0,1\}^{n}$, and the predicted labels by $\mathbf{\hat{y}}\in \{0,1\}^{n}$. $\mathbf{M}\in \{0,1\}^{n\times m}$ denotes a matching matrix, where $M_{i,j}$ indicates whether design $i$ matches template $j$ (1 for match, and 0 for not match). Binary vector $\mathbf{c}\in \{0,1\}^{m}$ denotes the chosen \Prototype{s}, where $c_j$ indicates whether template $j$ is chosen (1 for choose, and 0 for not choose). We solve the following ILP for template selection.

\begin{align}
    & \max \sum_{i=1}^{n} \hat{y}_i \text{\ \ \ \ \ (optimizing support)}  \label{ilp_obj}\\
    & \text{ s.t.} \sum_{j=1}^{m} c_j \leq s \text{\ \ \ \ \ (sparsity constraint)} \label{ilp_c_1}\\
    & \text{\ \ \ \ \ } \sum_{i=1}^{n} y_i \cdot \hat{y}_i \geq \left(\sum_{i=1}^{n} \hat{y}_i\right) \cdot p \text{\ \ (minimum precision)} \label{ilp_c_2}\\
    & \text{\ \ \ \ \ } \sum_{j=1}^{m} M_{i,j}\cdot c_j \geq \hat{y}_i,\ \ i=1 ,\ldots, n \text{\ \ \ (define $\hat{y}_i$)} \label{ilp_c_3}\\
    & \text{\ \ \ \ \ } \sum_{j=1}^{m} M_{i,j}\cdot c_j \leq m\cdot \hat{y}_i,\ \ i=1 ,\ldots, n \text{\ \ \ (define $\hat{y}_i$)} \label{ilp_c_4}\\
    & \text{\ \ \ \ \ } c_j\in\{0,1\}, \quad j=1 ,\ldots, m \\
    & \text{\ \ \ \ \ } \hat{y}_i\in\{0,1\}, \quad i=1 ,\ldots, n.
\end{align}

In this ILP, the objective (\ref{ilp_obj}) means maximizing the total number of designs predicted as positive, which is the same as support. Constraint (\ref{ilp_c_1}) controls the sparsity, i.e., choose at most $s$ \Prototype{s}. Constraint (\ref{ilp_c_2}) guarantees training precision of the model is at least $p$. (\ref{ilp_c_3}) and (\ref{ilp_c_4}) constraints together define $\hat{y}_i$, where $\hat{y}_i=1$ if and only if design $i$ matches at least one of the chosen \Prototype{s}. In particular, (\ref{ilp_c_3}) says that if design $i$ does not match any chosen template $j$ (i.e., whenever $c_j$ is 1, $M_{i,j}$ happens to be 0), then $\hat{y}_i$ will be set to 0. (\ref{ilp_c_4}) will ensure that if there is a match for design $i$ to any of the chosen templates (which all have $c_j=1$), then this design is assigned $\hat{y}_i$=1.
Using a commercial MIP solver, a problem with around 10000 candidate templates can be solved to optimality (when the current best solution meets the upper bound of the best possible solution) or near-optimality in about 10 minutes (running single-threaded on one core of a 2.66GHz Intel E5640 Xeon Processor). If $s=5$, it can be solved to optimality all the time, and when $s=10$, the optimality gap (difference between current best solution and an upper bound of the best possible solution as the percentage of the upper bound) is always $<$20\% for a run time of 30 minutes (running under the same environment); it is worthwhile to note that optimal solutions are often attained quickly, but the solvers can take a while to prove that the solution is optimal. Note that, if desired, one can set higher minimum precision and support thresholds for the pre-selection step to make the problem even smaller, so that the ILP can be solved even faster. We choose $s=5$ for all the experiments in the main paper.

The result of the ILP is our \PrototypeModel{}.

\subsubsection{Property-to-structure Sampling}
\label{sec:prototype:p2d}
After training the structure-to-property model, the resulting \PrototypeModel{} can be directly used to solve the inverse property-to-structure problem. An easy sampling procedure to do this is as follows: first, randomly choose a \Prototype{} $\mathbf{t}$ from the \PrototypeModel{}, where the probability to choose each template is proportional to its support; second, for all entries in $\mathbf{t}$ that equal *, randomly assign value 0 or 1 to them. These sampled unit-cells are likely to have the desired band gap.

\subsubsection{Transfer to Finer Resolution}
\label{sec:prototype:transfer}
The \PrototypeModel{} naturally transfers coarse scale information to finer resolutions. In particular, by subdividing each pixel in the \Prototype{} into four sub-pixels, we directly obtain a \Prototype{} defined on a finer-resolution space. 

Algorithm \ref{alg:template} shows the entire pipeline of using \PrototypeModel{} to sample fine resolution designs with the target band gap property. Visual illustration of the sampling process can be found in Supplementary Information C.

\begin{algorithm*}[ht]
 \caption{Sampling Fine Resolution Unit-cell Designs via Unit-cell Template Set}
 \begin{tabbing}
 xxx \= xx \= xx \= xx \= xx \= xx \kill
 \textbf{Input}: simulated coarse resolution ($10\times 10$) dataset $\mathcal{D}:=\{\mathbf{x}_i,y_i\}_{i=1}^{2^{15}}$, $\mathbf{x}_i\in \{0,1\}^{15}$: raw features; $y_i$: band gap label\\
 \textbf{Parameters:} pre-selection support $\psi_{\text{pre}}$, pre-selection precision $p_{\text{pre}}$, sparsity constraint $s$, minimum precision $p$ \\
 \textbf{Output:} raw features of a fine resolution unit-cell $\mathbf{\tilde{x}}$\\
 1: \> pre-select candidate template set $S_T^{\text{pre}}$ = pre-selecting($\mathcal{D}$, $\{0,1,*\}^{15}$, $\psi_{\text{pre}}$, $p_{\text{pre}}$), see Section \ref{sec:preselect}\\
 2: \> run ILP to find optimal template set $S_T^{*}$ = ILP($\mathcal{D}$, $S_T^{\text{pre}}$, $s$, $p$), see Section \ref{sec:ilp}\\
 3: \> randomly pick a template $\mathbf{t}\in S_T^{*}$ \\
 4: \> expand $\mathbf{t}$ to fine resolution space, get $\mathbf{\tilde{t}}$\\
 5: \> randomly set to 0 or 1 for all * elements in $\mathbf{\tilde{t}}$, get $\mathbf{\tilde{x}}$\\
 6: \> \textbf{return} $\mathbf{\tilde{x}}$
 \end{tabbing}
 \label{alg:template}
\end{algorithm*}

\section{Results}
\label{sec:experiments}
The results are organized according to four objectives we want to achieve with the proposed methods, including (a) design-to-property prediction; (b) property-to design sampling; (c) identify key patterns; (d) transfer to finer resolution. We evaluate how well the proposed methods can achieve these objectives, followed by several tests involving practical applications in materials discovery.
\begin{table*}
\begin{subtable}{1\textwidth}
\centering
\begin{tabular}{|c|c|c|c|c|c|c|}
\hline
\multicolumn{2}{|c|}{\multirow{2}{*}{Model}} & \multicolumn{5}{c|}{Frequency range} \\ \cline{3-7}
\multicolumn{2}{|c|}{} &\lbrack0, 10\rbrack kHz& \lbrack10, 20\rbrack kHz & \lbrack20, 30\rbrack kHz & \lbrack30, 40\rbrack kHz & \lbrack40, 50\rbrack kHz \\ \hline 
\multirow{3}{*}{SVM} & raw & 71.77\% & 73.88\% & 50.85\% & 54.93\% & 49.84\% \\ \cline{2-7}
 & SFF (Ours)& 75.62\% & 77.35\% & 67.96\% & 59.89\% & 49.93\% \\ \cline{2-7}
 & improvement & +3.85\% & +3.47\% & +17.11\% & +4.96\% & 0.09\% \\\hline
\multirow{3}{*}{LR} & raw & 78.03\% & 75.44\% & 56.31\% & 76.59\% & 90.96\% \\ \cline{2-7}
 & SFF  (Ours)& 80.53\% & 79.55\% & 69.04\% & 76.9\% &  91.32\% \\ \cline{2-7}
 & improvement & +2.50\% & +4.11\% & +12.73\% & +0.31\% &  +0.36\% \\ \hline
\multirow{3}{*}{RF} & raw & 85.81\% & 80.04\% & 73.00\% & 77.66\% & 87.54\% \\ \cline{2-7}
 & SFF  (Ours)& 85.52\% & 81.98\% & 74.53\% & 81.26\% & \textbf{95.35\%} \\ \cline{2-7}
 & improvement & -0.29\% & +1.94\% & +1.53\% & +3.60\% & +7.81\%\\ \hline
\multirow{3}{*}{CART} & raw & 84.74\% & 75.68\% & 63.40\% &  73.95\% & 86.82\% \\ \cline{2-7}
 & SFF (Ours)& 83.63\% & 80.13\% & 70.72\% & 79.73\% & 94.47\% \\ \cline{2-7}
 & improvement & -1.11\% & +4.45\% & +7.32\% & +5.78\% & +7.65\% \\ \hline
\multirow{3}{*}{MLP} & raw & 90.72\% & 86.39\% & 78.15\% & 77.47\% & 65.90\% \\ \cline{2-7}
 & SFF (Ours)& 85.90\% & 82.55\% & 76.01\% & 77.69\% & 66.4\%\\ \cline{2-7}
 & improvement & -4.82\% & -3.84\% & -2.14\% & +0.22\% & +0.50\%\\ \hline
\multirow{3}{*}{LightGBM} & raw & 91.32\% & 88.27\% & 81.11\% & 82.62\% & 77.76\% \\ \cline{2-7}
 & SFF (Ours) & \textbf{96.10\%} & \textbf{90.97\%} & \textbf{85.57\%} & \textbf{90.01\%} & 94.76\% \\\cline{2-7}
 & improvement & +4.78\% & +2.70\% & +4.46\% & +7.39\% & +17.00\% \\ \hline
CNN & raw & 93.24\% & 89.76\% & 81.65\% & 79.56\% & 84.83\% \\ \hline
\end{tabular}
\subcaption{structure-to-property prediction: testing balanced accuracies (baccs) of different methods. We mark the models with the best bacc in each frequency range in bold. }
\vspace{.1in}
\end{subtable}
\begin{subtable}{1\textwidth}
\centering
\begin{tabular}{|c|c|c|c|c|}
        \hline
         Frequency range & Raw+LightGBM & SFF+LightGBM & SFF+GOSDT  (Ours) & Unit-cell Template Sets (Ours) \\\hline
         \lbrack0, 10\rbrack kHz & 80.77\%, 2310 & 88.93\%, 2169  & 95.77\%, 89& \textbf{98.53\%}, 339\\  \hline
         \lbrack10, 20\rbrack kHz & 93.52\%, 4013 & 95.62\%, 4063 & \textbf{98.11\%}, 423 & \textbf{98.68\%}, 758\\\hline
         \lbrack20, 30\rbrack kHz & 86.94\%, 3654 & 89.56\%, 3811 & \textbf{94.15\%}, 205 & \textbf{94.08\%}, 203\\\hline
    \end{tabular}
    \subcaption{Property-to-structure sampling: testing precision and support of different methods. Numbers in the table cells are formatted as ``precision, support.'' The testing support here is calculated among 6554 testing samples (20\% of the entire dataset).}
\vspace{.1in}
\end{subtable}
\begin{subtable}{1\textwidth}
\centering
\begin{tabular}{|c|c|c|c|c|c|c|}
        \hline
         \multirow{2}{*}{Frequency range} & CNN+resizing & LightGBM+resizing & SFF+GOSDT (Ours)& \multicolumn{3}{c|}{Unit-cell Template Sets (Ours)} \\ \cline{2-7}
         & $20\times 20$ & $20\times 20$ & $20\times 20$ & $20\times 20$ & $40\times 40$ & $80\times 80$\\\hline
         \lbrack0, 10\rbrack kHz & 18.0\% & 25.0\% & 72.5\% & \textbf{100.0\%} & \textbf{100.0\%} & \textbf{100.0\%} \\  \hline
         \lbrack10, 20\rbrack kHz & 58.0\% & 68.5\% & 73.5\% & \textbf{98.5\%} & \textbf{99.0\%} & \textbf{100.0\%} \\\hline
         \lbrack20, 30\rbrack kHz & 39.5\% & 52.5\% & 25.0\% & \textbf{91.0\%} & \textbf{96.0\%} & \textbf{98.0\%}\\\hline
    \end{tabular}
    \subcaption{Transfer to finer resolution: transfer precision of different methods.}
\end{subtable}
\caption{Summary of key quantitative results. The results are organized with respect to different objectives of data-driven metamaterials design. (a) structure-to-property prediction; (b) Property-to-structure sampling; (c) Transfer to finer resolution.} \label{tab:summary}
\end{table*}
\subsection{Objective 1: Structure-to-property Prediction}
\label{sec:experiments:prediction}
Here, we test how well the proposed methods can perform structure-to-property prediction. We chose five frequency ranges ([0, 10], [10, 20], [20, 30], [30, 40] and [40, 50] kHz) to predict the existence of band gaps; these frequency ranges correspond to five different binary classification problems. Using balanced accuracy (bacc) as the evaluation metric, we compare the predictive performance of a diverse set of ML models with and without the shape-frequency features. We specifically consider linear models like support vector machines with linear kernels (SVMs) and logistic regression (LR); tree-based models like CART, random forest (RF), and boosted trees (LightGBM \citep{ke2017lightgbm}), as well as neural networks including the multi-layer perceptron (MLP). We also compare the proposed method with convolutional neural networks (CNNs), since they have been widely used in previous works of ML-based metamaterial design. The baccs of each model trained on raw feature, SFF, and improvements of SFF over raw features, are shown in Table \ref{tab:summary} (a). We train each model 5 times and average the accuracy.

Our results show that \textbf{using the shape-frequency features, rather than the original raw features, improves the accuracy of classifiers for most machine learning methods}, especially tree-based methods such as boosted trees, but with the exception of MLP (SFF decreases its in [0, 10] kHz, [10, 20] kHz, and [20, 30] kHz).

One might expect CNNs to achieve great success in classifying band gaps for 2-D metamaterials since the unit-cells share many similarities with images. However, LightGBM \citep{ke2017lightgbm} built on shape-frequency features outperforms  ResNet18 \citep{he2016deep} in all ranges. In some cases, e.g., within frequency range [40, 50] kHz, simple models like CART outperform CNNs.

More details of the experiment (e.g., hyper-parameter settings) can be found in Supplementary Information D.1.

\subsection{Objective 2: Property-to-structure Sampling}
Using the methods discussed in Section \ref{sec:sff:p2d} and Section \ref{sec:prototype:p2d}, we are able to solve the inverse design (property-to-structure sampling) problem.

In practice, materials scientists need valid designs with the target property, but they do not require the set of \textit{all} designs with the property. 
As such, our performance metric is precision, rather than recall.
We also calculate the support, which is the total number of testing samples predicted as positive, to ensure the models can generate enough potentially-valid designs. 
Table \ref{tab:summary} (b) lists the precision and support values from different methods. 
The methods we compared include GOSDT trained on shape-frequency features (denoted SFF) with the objective in Section \ref{sec:gosdt_objective}, the \PrototypeModel{}, and LightGBMs trained on SFF and raw features.

In terms of precision, \textbf{SFF+GOSDT and \PrototypeModel{s} significantly outperformed LightGBMs}.  This is probably owing to the fact that the proposed methods directly optimize precision. LightGBMs maintain larger support, while the support of SFF+GOSDT and \PrototypeModel{s} is much lower. But for practical use, it is sufficient that the model finds dozens of valid designs. The average sampling time of these methods and the results of \PrototypeModel{s} with different sparsity constraints can be found in Supplementary Information D.2.

\subsection{Objective 3: Show Key Patterns}
\label{sec:experiments:key_patterns}
One advantage of the proposed methods is model interpretability; we aim to explicitly identify the key patterns learned from data that are related to the target property. In this way, domain experts can verify whether the learned rules are aligned with the domain knowledge, or even discover new knowledge. 

In Figure \ref{fig:shapes}\textbf{b} and Figure \ref{fig:unitcell_template_set}\textbf{b}, we visualize the GOSDT+SFF and \PrototypeModel{} learned for band gaps in several frequency ranges ([0, 6], [6, 12], [12, 18], [18, 24], [24, 30] kHz).

In Figure \ref{fig:shapes}\textbf{b}, the top splits of each tree trained on shape-frequency features seem to be looking for bars in the soft material. Taking the top-right tree in \ref{fig:shapes}\textbf{b} as an example, the root node checks if the $1\times 4$ soft bar occurs in more than 22\% of the places in the unit-cell. The unit-cell frequencies need to pass the thresholds to get to the ``band gap'' node.
All band gap nodes are on the left branch of the deepest decision nodes in the trees. For instance, the deepest decision node of the top-right tree checks if shape 17, two $1\times 4$ soft bars 2 pixels away from each other, occurs with more than 5\% frequency in the unit-cell. To reach the band gap prediction node, the sample needs to go to the left branch, where shape 17 should occur with less than 5\% frequency. This indicates the unit-cells should not have too many soft material patterns. In the field of elastic wave propagation, it is known that the presence of stiff inclusions in a matrix of a softer material may open a band gap due to scattering or resonant dynamics. The trees we found seem to be looking for patterns that fit this description: the deepest node encourages the existence of stiff inclusion while the first node encourages more soft material. 

The unit-cell templates (Figure \ref{fig:unitcell_template_set}\textbf{b}) 
contrast with the shape frequency features in that they explicitly identify global (rather than local) patterns. In the unit-cell template sets for different frequency ranges, we can observe the existence of soft circles (closed curves) and stiff inclusions inside the circles, which are responsible for the formation of band gaps. As the frequency range moves higher, the size of the circle decreases. This supports the physical intuition that the smaller the stiff inclusions, the higher the frequency of the band gap.

\subsection{Objective 4: Transfer to Finer Resolution}

In Sections \ref{sec:sff:transfer} and \ref{sec:prototype:transfer}, we discussed how the proposed methods can transfer coarse scale information to finer resolution design space. To evaluate how well the model can transfer information, we trained the models on coarse resolution ($10\times 10$) unit-cells and tested them on finer resolution ($20\times 20$, $40\times 40$ and $80\times 80$) unit-cells. Table \ref{tab:summary}(c) shows the transfer precision of GOSDT+SFF and \PrototypeModel{s} for band gaps in different frequency ranges. Other ML methods are not directly comparable because standard ML models trained on 10$\times$10 data cannot take in 20$\times$20 data. Therefore, we compared the proposed methods with baselines with slight modifications: we resized the fine resolution (20$\times$20) unit-cell to the original size (10$\times$10) and applied two algorithms (CNN or LightGBM) for rejection sampling. As before, if the CNN or LightGBM model predicts that the resized design has a band gap, we accept that sample, otherwise we reject it. The resizing was done via bicubic interpolation. Here, we did not compare with deep generative models such as GANs. Although GANs (which are notoriously hard to train) might generate materials faster than rejection sampling, their precision can only be lower because GANs' discriminators are co-trained with the generator, and thus cannot be more accurate than a CNN directly trained to predict only the target.
For each frequency range, we asked the trained models to sample 200 unit-cell designs in finer resolution space, and ran the FEA simulation to obtain the true band gap property for evaluating the transfer precision. 
For the baseline models and GOSDT+SFF, we show the results for $20\times 20$ design space. As \PrototypeModel{s} performs extremely well on this task, and generates new designs efficiently, we also show its results in $40\times 40$ and $80\times 80$ design space. Please see Supplementary Information C for visual illustrations of how to sample fine resolution designs using each model.

The results in Table \ref{tab:summary}(c) indicate that \textbf{the \Prototype{s}, when transferred to all finer resolutions ($20\times 20$, $40\times 40$ or $80\times 80$), have very high precision, with almost no precision drop compared to $10\times 10$}. GOSDT+SFF and other baselines do not generalize as well as \Prototype{s} to the finer resolution design space. GOSDT+SFF performs better than the resizing baselines in [0, 10] kHz and [10, 20] kHz, but performs worse than baselines in [20, 30] kHz. Interestingly, the transfer precisions, of both GOSDT+SFF and \PrototypeModel{s}, decrease as frequency ranges moves higher, although \PrototypeModel{s} have a much slower precision decrease than GOSDT+SFF. A possible explanation for the decrease of precision is that, in higher frequency ranges, the band gaps are physically more related to finer scale features that are not included in the coarse resolution dataset. Since the models are trained on coarse resolution data, they can only transfer physics that occurs in coarse patterns to finer resolution design space, but cannot discover finer scale physics without supervision. But as shown by the transfer precision results, we should emphasize that our \PrototypeModel{s} method was capable of extracting critical coarse resolution features such that this decrease of precision at finer resolution due to wave physics is minimized (the worst transfer precision is still above 90\%). One further potential improvement to this is to add new samples at each finer resolution design space when transferring between extreme scales.

In Supplementary Information D.3, we show additional results on sampling with correlation between green pixels for \PrototypeModel{s}, which demonstrates the surprising flexibility of \PrototypeModel{s} in terms of designing at finer-resolution design space. 

\subsection{Practicality Test}

\begin{figure*}
    \centering
    \includegraphics[width=6.3in]{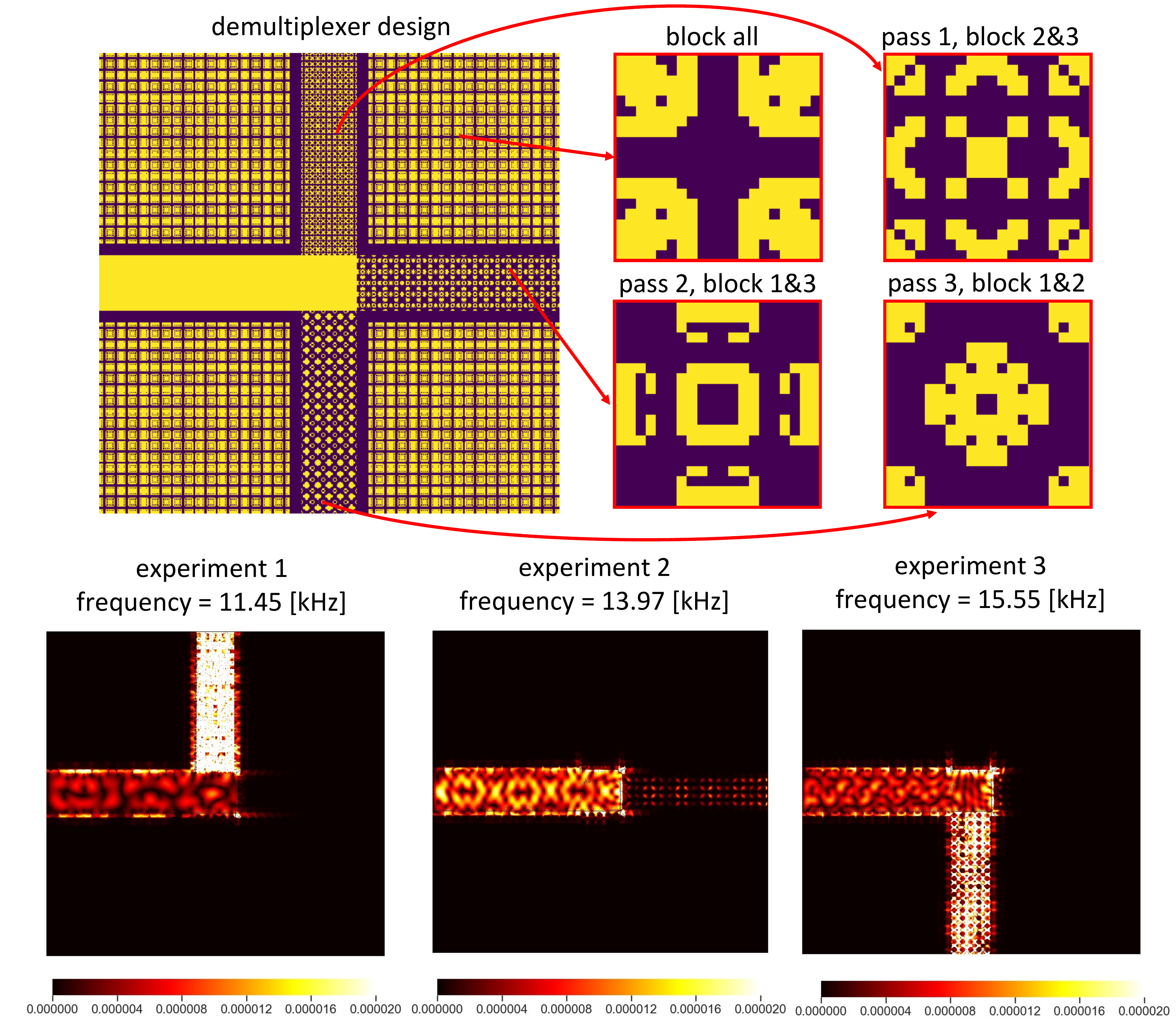}
    \caption{Mechanical wave demultiplexer and the displacement fields for input signals frequencies. Each part of the demultiplexer  was built using our discovered unit-cells that have desired properties. \textit{Top}: The demultiplexer and the unit-cell designs used to make the demultiplexer. \textit{Bottom}: Magnitude of displacement fields when signals of different frequencies (11.45, 13.97 and 15.55 [kHz]) are fed into the left side of the demultiplexer. The displacement values are clipped if they exceed the display range. The signals with different frequencies go through different channels in the demultiplexer: 11.45 [kHz] goes upward, 13.97 [kHz] goes right, and 15.55 [kHz] goes downward, as desired. Note that, although the passing signal of experiment 2 is not as strong as signals in the other experiments, it still pass the channel on the right without fading inside the channel.}
    \label{fig:demultiplexer}
\end{figure*}

In our method and simulations so far, we assumed the unit-cell is tiled infinitely for computational convenience. However, a unit-cell can only be tiled finitely in practice and the results can differ for infinite and finitely tiled domains due to boundary conditions. To test whether the designs found by our method work in practice, we simulated the dispersion relations of finitely-tiled materials made by unit-cell designs discovered by our method. See Supplementary Information D.4 for results of the finite tiling COMSOL simulation. The results show that our method is robust under finite tiling.

In addition to the finite tiling test, we also tested the practicality of the proposed method on its ability to create a wave demultiplexer (Figure \ref{fig:demultiplexer}), in which waves with different frequencies travel through the materials in different directions. That is, a signal enters the demultiplexer, and there are three different possible outputs; which one will be non-zero  depends on the frequency of the input signal. Specifically, we will build the demultiplexer to route signals from three different frequency ranges in different directions. 

We need 5 different materials to build the demultiplexer: one material with band gaps covering all three ranges, one material allowing band pass in all ranges, and three materials allowing band pass in one range while blocking the other two ranges. We use homogeneous stiff unit-cells for the material allowing band pass in all ranges. For other materials, we train a \PrototypeModel{} to find $20\times 20$ unit-cell designs with these properties, and assemble the 5 unit-cells to build the demultiplexer (top row of Figure \ref{fig:demultiplexer}). The bottom row of Figure \ref{fig:demultiplexer} shows the how the demultiplexer successfully guides waves with different frequencies (11.45, 13.97 and 15.55 [kHz]) towards different directions, which was the goal of the experiment.

\section{Conclusion and Discussion}
\label{sec:conclusion}
Our work shows the power of interpretable machine learning tools in material design. The approach has achieved both mechanistic understanding, e.g., physically interpretable rules of patterns that lead to band gaps, and designing new materials with desired functionality. The approach has been demonstrated to be predictive for both infinite domains and realistic finite domains, and it has been able to design the material geometry for a wave demultiplexer. Additionally, our multi-resolution framework, which robustly carries coarse-scale knowledge to finer resolutions, is potentially applicable to a wide range of materials science problems.

Since our method learns robust coarse-scale features that can generalize to finer-resolution design space, it might also be useful in future studies for determining how to collect finer-resolution training data for rapidly capturing fine-scale physics. Using these new data, we might be able to fine tune the model so that it can more efficiently capture physics at multiple scales.


\section*{Code and Data Availability}
The code and data for replicating our results are available on \url{https://github.com/zhiCHEN96/interpretable_ml_metamaterials.git}.

\bibliographystyle{plain}
\bibliography{ref.bib}

\begin{thebibliography}{10}

\bibitem{38}
K.~Bertoldi, V.~Vitelli, J.~Christensen, and M.~van Hecke.
\newblock Flexible mechanical metamaterials.
\newblock {\em Nature Reviews Materials}, 2:17066, 2017.

\bibitem{bien2011prototype}
Jacob Bien and Robert Tibshirani.
\newblock {Prototype Selection for Interpretable Classification}.
\newblock {\em The Annals of Applied Statistics}, pages 2403--2424, 2011.

\bibitem{Bilal_PRE_2011}
O.~R. Bilal and M.~I. Hussein.
\newblock Ultrawide phononic band gap for combined in-plane and out-of-plane
  waves.
\newblock {\em Physical Review E}, 84:065701, 2011.

\bibitem{bostanabad2019globally}
Ramin Bostanabad, Yu-Chin Chan, Liwei Wang, Ping Zhu, and Wei Chen.
\newblock Globally approximate gaussian processes for big data with application
  to data-driven metamaterials design.
\newblock {\em Journal of Mechanical Design}, 141(11), 2019.

\bibitem{clark1991rule}
Peter Clark and Robin Boswell.
\newblock Rule induction with {CN2}: Some recent improvements.
\newblock In {\em European Working Session on Learning}, pages 151--163.
  Springer, 1991.

\bibitem{creswell2018generative}
Antonia Creswell, Tom White, Vincent Dumoulin, Kai Arulkumaran, Biswa Sengupta,
  and Anil~A Bharath.
\newblock Generative adversarial networks: An overview.
\newblock {\em IEEE Signal Processing Magazine}, 35(1):53--65, 2018.

\bibitem{deng2021neural}
Yang Deng, Simiao Ren, Kebin Fan, Jordan~M Malof, and Willie~J Padilla.
\newblock Neural-adjoint method for the inverse design of all-dielectric
  metasurfaces.
\newblock {\em Optics Express}, 29(5):7526--7534, 2021.

\bibitem{elzouka2020interpretable}
Mahmoud Elzouka, Charles Yang, Adrian Albert, Sean Lubner, and Ravi~S Prasher.
\newblock Interpretable inverse design of particle spectral emissivity using
  machine learning.
\newblock {\em arXiv preprint arXiv:2002.04223}, 2020.

\bibitem{finol2019deep}
David Finol, Yan Lu, Vijay Mahadevan, and Ankit Srivastava.
\newblock Deep convolutional neural networks for eigenvalue problems in
  mechanics.
\newblock {\em International Journal for Numerical Methods in Engineering},
  118(5):258--275, 2019.

\bibitem{frank1998generating}
Eibe Frank and Ian~H Witten.
\newblock Generating accurate rule sets without global optimization.
\newblock In {\em Proceedings of International Conference on Machine Learning
  ({ICML})}, pages 144--151, 1998.

\bibitem{goodfellow2014generative}
Ian Goodfellow, Jean Pouget-Abadie, Mehdi Mirza, Bing Xu, David Warde-Farley,
  Sherjil Ozair, Aaron Courville, and Yoshua Bengio.
\newblock Generative adversarial nets.
\newblock {\em Advances in neural information processing systems}, 27, 2014.

\bibitem{he2016deep}
Kaiming He, Xiangyu Zhang, Shaoqing Ren, and Jian Sun.
\newblock Deep residual learning for image recognition.
\newblock In {\em Proceedings of the IEEE Conference on Computer Vision and
  Pattern Recognition}, pages 770--778, 2016.

\bibitem{jiang2020deep}
Jiaqi Jiang, Mingkun Chen, and Jonathan~A Fan.
\newblock Deep neural networks for the evaluation and design of photonic
  devices.
\newblock {\em Nature Reviews Materials}, pages 1--22, 2020.

\bibitem{ke2017lightgbm}
Guolin Ke, Qi~Meng, Thomas Finley, Taifeng Wang, Wei Chen, Weidong Ma, Qiwei
  Ye, and Tie-Yan Liu.
\newblock Lightgbm: A highly efficient gradient boosting decision tree.
\newblock In {\em Advances in Neural Information Processing Systems}, pages
  3146--3154, 2017.

\bibitem{kim2014bayesian}
Been Kim, Cynthia Rudin, and Julie~A Shah.
\newblock {The Bayesian Case Model: A Generative Approach for Case-Based
  Reasoning and Prototype Classification}.
\newblock In {\em Proceedings of Neural Information Processing Systems
  {(NeurIPS)}}, volume~27, pages 1952--1960, 2014.

\bibitem{krizhevsky2012imagenet}
Alex Krizhevsky, Ilya Sutskever, and Geoffrey~E Hinton.
\newblock Imagenet classification with deep convolutional neural networks.
\newblock {\em Advances in Neural Information Processing Systems},
  25:1097--1105, 2012.

\bibitem{lakkaraju2016interpretable}
Himabindu Lakkaraju, Stephen~H Bach, and Jure Leskovec.
\newblock Interpretable decision sets: A joint framework for description and
  prediction.
\newblock In {\em Proceedings of the 22nd ACM SIGKDD International Conference
  on Knowledge Discovery and Data Mining}, pages 1675--1684, 2016.

\bibitem{GOSDT}
Jimmy Lin, Chudi Zhong, Diane Hu, Cynthia Rudin, and Margo Selzer.
\newblock Generalized optimal sparse decision trees.
\newblock In {\em Proc. International Conference on Machine Learning}, 2020.

\bibitem{liu2020compounding}
Zhaocheng Liu, Dayu Zhu, Kyu-Tae Lee, Andrew~S Kim, Lakshmi Raju, and Wenshan
  Cai.
\newblock Compounding meta-atoms into metamolecules with hybrid artificial
  intelligence techniques.
\newblock {\em Advanced Materials}, 32(6):1904790, 2020.

\bibitem{liu2018generative}
Zhaocheng Liu, Dayu Zhu, Sean~P Rodrigues, Kyu-Tae Lee, and Wenshan Cai.
\newblock Generative model for the inverse design of metasurfaces.
\newblock {\em Nano Letters}, 18(10):6570--6576, 2018.

\bibitem{ma2018deep}
Wei Ma, Feng Cheng, and Yongmin Liu.
\newblock Deep-learning-enabled on-demand design of chiral metamaterials.
\newblock {\em ACS nano}, 12(6):6326--6334, 2018.

\bibitem{ma2019probabilistic}
Wei Ma, Feng Cheng, Yihao Xu, Qinlong Wen, and Yongmin Liu.
\newblock Probabilistic representation and inverse design of metamaterials
  based on a deep generative model with semi-supervised learning strategy.
\newblock {\em Advanced Materials}, 31(35):1901111, 2019.

\bibitem{ma2020data}
Wei Ma and Yongmin Liu.
\newblock A data-efficient self-supervised deep learning model for design and
  characterization of nanophotonic structures.
\newblock {\em SCIENCE CHINA Physics, Mechanics \& Astronomy}, 63(8):1--8,
  2020.

\bibitem{ma2021deep}
Wei Ma, Zhaocheng Liu, Zhaxylyk~A Kudyshev, Alexandra Boltasseva, Wenshan Cai,
  and Yongmin Liu.
\newblock Deep learning for the design of photonic structures.
\newblock {\em Nature Photonics}, 15(2):77--90, 2021.

\bibitem{multiphysics1998introduction}
{COMSOL} Multiphysics.
\newblock Introduction to {COMSOL} multiphysics{\textregistered}.
\newblock {\em {COMSOL} Multiphysics, Burlington, MA, accessed Feb}, 9:2018,
  1998.

\bibitem{nadell2019deep}
Christian~C Nadell, Bohao Huang, Jordan~M Malof, and Willie~J Padilla.
\newblock Deep learning for accelerated all-dielectric metasurface design.
\newblock {\em Optics express}, 27(20):27523--27535, 2019.

\bibitem{qiu2019deep}
Tianshuo Qiu, Xin Shi, Jiafu Wang, Yongfeng Li, Shaobo Qu, Qiang Cheng, Tiejun
  Cui, and Sai Sui.
\newblock Deep learning: a rapid and efficient route to automatic metasurface
  design.
\newblock {\em Advanced Science}, 6(12):1900128, 2019.

\bibitem{ren2020three}
Haoran Ren, Wei Shao, Yi~Li, Flora Salim, and Min Gu.
\newblock Three-dimensional vectorial holography based on machine learning
  inverse design.
\newblock {\em Science advances}, 6(16):eaaz4261, 2020.

\bibitem{39}
C.~Schumacher, B.~Bickel, J.~Rys, S.~Marschner, C.~Daraio, and M.~Gross.
\newblock Microstructures to control elasticity in 3d printing.
\newblock {\em ACM Transactions on Graphics (TOG)}, 34:136, 2015.

\bibitem{37}
O.~Sigmund.
\newblock In systematic design of metamaterials by topology optimization.
\newblock In Iutam Symposium, editor, {\em on Modelling Nanomaterials and
  Nanosystems, Dordrecht, 2009; Pyrz, R}, pages 151--159. Dordrecht, Rauhe, J.
  C., Eds. Springer Netherlands, 2009.

\bibitem{sigmund2003systematic}
Ole Sigmund and Jakob S{\o}ndergaard~Jensen.
\newblock Systematic design of phononic band--gap materials and structures by
  topology optimization.
\newblock {\em Philosophical Transactions of the Royal Society of London.
  Series A: Mathematical, Physical and Engineering Sciences},
  361(1806):1001--1019, 2003.

\bibitem{wang2020deep}
Liwei Wang, Yu-Chin Chan, Faez Ahmed, Zhao Liu, Ping Zhu, and Wei Chen.
\newblock Deep generative modeling for mechanistic-based learning and design of
  metamaterial systems.
\newblock {\em Computer Methods in Applied Mechanics and Engineering},
  372:113377, 2020.

\bibitem{WangEtAl16}
Tong Wang, Cynthia Rudin, Finale Doshi, Yimin Liu, Erica Klampfl, and Perry
  MacNeille.
\newblock Bayesian {Or}'s of {And}'s for interpretable classification with
  application to context aware recommender systems.
\newblock In {\em International Conference on Data Mining {(ICDM)}}, 2016.

\bibitem{WangEtAl2017}
Tong Wang, Cynthia Rudin, Finale Doshi-Velez, Yimin Liu, Erica Klampfl, and
  Perry MacNeille.
\newblock A bayesian framework for learning rule sets for interpretable
  classification.
\newblock {\em Journal of Machine Learning Research}, 18(70):1--37, 2017.

\bibitem{wiecha2019deep}
Peter~R Wiecha and Otto~L Muskens.
\newblock Deep learning meets nanophotonics: a generalized accurate predictor
  for near fields and far fields of arbitrary 3d nanostructures.
\newblock {\em Nano letters}, 20(1):329--338, 2019.

\bibitem{xu2021interfacing}
Yihao Xu, Xianzhe Zhang, Yun Fu, and Yongmin Liu.
\newblock Interfacing photonics with artificial intelligence: an innovative
  design strategy for photonic structures and devices based on artificial
  neural networks.
\newblock {\em Photonics Research}, 9(4):B135--B152, 2021.

\bibitem{zhu2022harnessing}
Yi~Zhu and Evgueni~T Filipov.
\newblock Harnessing interpretable machine learning for origami feature design
  and pattern selection.
\newblock {\em arXiv preprint arXiv:2204.07235}, 2022.

\bibitem{44}
O.~C. Zienkiewicz, R.~L. Taylor, and J.~Z. Zhu.
\newblock {\em The Finite Element Method: Its Basis and Fundamentals (Seventh
  Edition)}.
\newblock Oxford, Butterworth-Heinemann, 2013.

\end{thebibliography}

\section*{Acknowledgements}
The authors are grateful to M. Bastawrous, A. Lin, K. Liu, C. Zhong, O. Bilal, W. Chen, C. Tomasi, and S. Mukherjee for the feedback and assistance they provided during the development and preparation of this research. The authors acknowledge funding from the National Science Foundation under grant OAC-1835782, Department of Energy under grants DE-SC0021358 and DE-SC0023194, and National Research Traineeship Program under NSF grants DGE-2022040 and CCF-1934964.

\section*{Author Contributions}
Z.C. developed the methods, designed metrics, designed visualization, and ran experiments related to the ML algorithms. A.O. developed the simulation code, generated the data, and did the practicality test. C.R., L.C.B., and C.D. conceived and supervised the project. All authors discussed the results and contributed to the writing.

\section*{Competing Interests}
The authors declare no competing interests.

\newpage
\appendix
\onecolumn
\renewcommand\refname{Supplementary References}
\renewcommand\thefigure{S\arabic{figure}}
\renewcommand\thetable{S\arabic{table}}
\setcounter{figure}{0}
\setcounter{table}{0}

\section{Simulation Settings} \label{app:simulsetup}

\subsection*{In-house/Infinite-medium Simulator}

To determine whether a unit-cell design has a bandgap, we evaluate the dispersion relation along the wavevector contour (Irreducible Brillouin Zone contour) shown in Supplementary Figure \ref{fig:wavevector}. To evaluate the dispersion relation, we solve a series of eigenvalue problems given by the harmonic elastic wave equation with Bloch-Floquet periodic boundary conditions.

Dispersion relation analysis is performed only on the domain of a single unit-cell, and assumes that the metamaterial has a unit cell that is infinitely repeating in each direction that Bloch-Floquet periodic boundary conditions are applied.

The harmonic elastic wave equation is:	

\begin{equation} \label{eq:harmonic_elastic_wave_equation_1}
    \rho(x) \omega^2 u = \frac{\partial}{\partial x_j} \left[ C_{ijkl}(x) \frac{1}{2} \left( \frac{\partial u_k}{\partial x_l} + \frac{\partial u_l}{\partial x_k} \right) \right] \quad i,j,k,l = 1,2.
\end{equation}

where $C$ is the material stiffness tensor, $\rho$ is the material density, $u$ is the displacement field (eigenfunction), $\omega$ is the eigenfrequency, and $x$ represents positions in the 2-D space. 
Displacements (and therefore wave modes) are confined to the same 2-D plane that contains the metamaterial. Bloch-Floquet theory says that the displacement field solutions, $u$, should be spatially periodic, so we introduce the wavevector $\gamma$ to characterize this spatial periodicity. To enforce Bloch-Floquet periodicity, we apply Bloch-Floquet periodic boundary conditions:



\begin{equation} \label{eq:harmonic_elastic_wave_equation_2}
u(x + a_n) = u(x) e^{-\mathbf{i} \gamma \cdot a_n} \ \forall x \in \{ x \in \Omega \ | \ x + a_n \in \Omega \}, \quad n = 1,2.
\end{equation}

where $\mathbf{i}$ is the imaginary unit, $a_n$ are the lattice vectors representing the periodicity of the material, and  $\Omega \subset \mathbb{R}^2$ represents the domain of the unit-cell.

To discretize and solve this eigenvalue problem, we apply the Finite Element method using bilinear quadrilateral elements. The resulting discretized eigenvalue problem is

\begin{equation}
\lbrack K(\gamma) - \omega^2 M(\gamma)\rbrack u = 0,
\end{equation} 


where $K$ and $M$ are the stiffness and mass matrices, respectively. The boundary conditions are baked into the wavevector-dependent stiffness and mass matrices. This code is implemented in \textit{MATLAB}.



\begin{figure}[H]
    \centering
    \includegraphics[width = .45\linewidth]{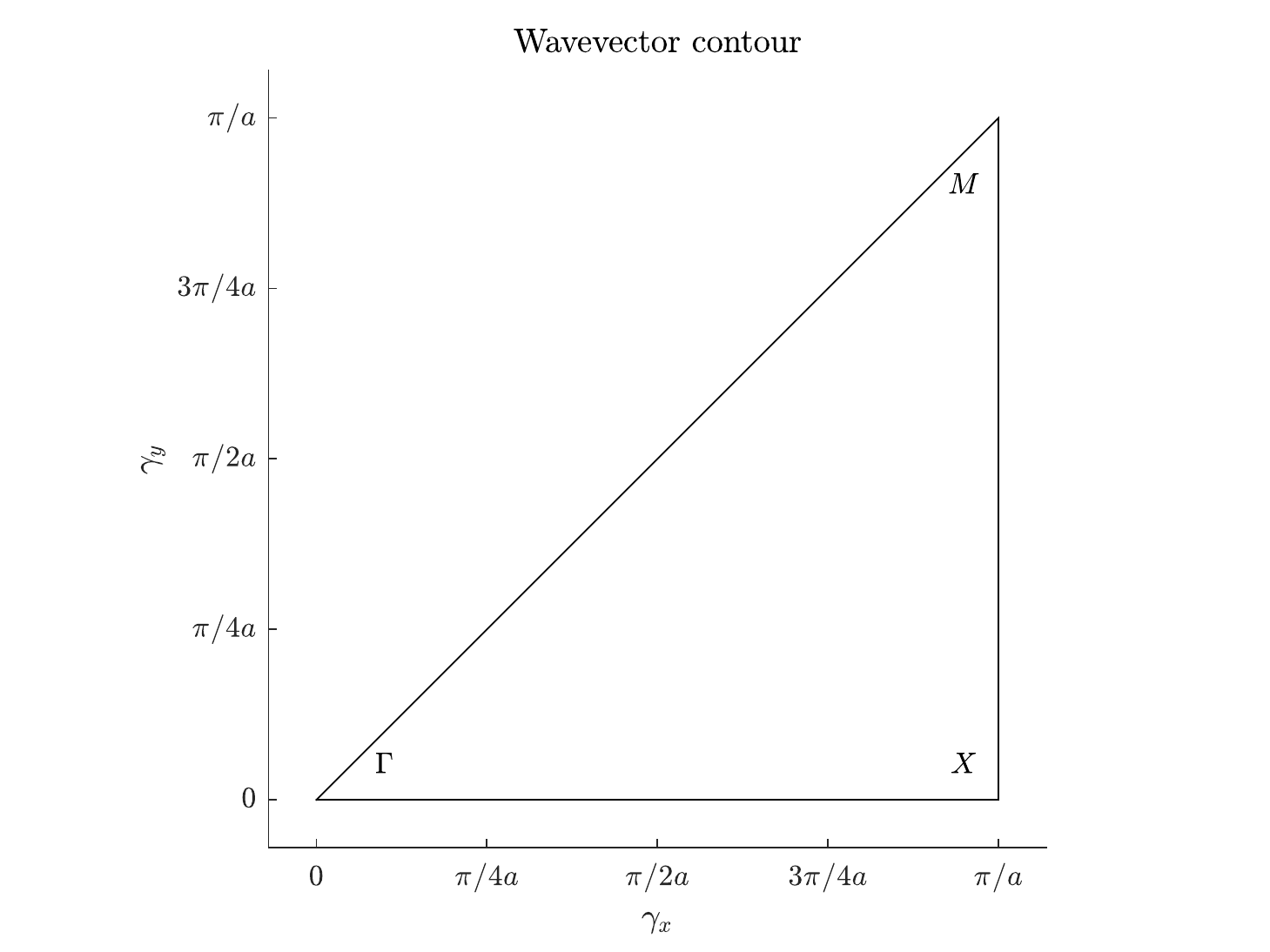}
    \includegraphics[width = .45\linewidth]{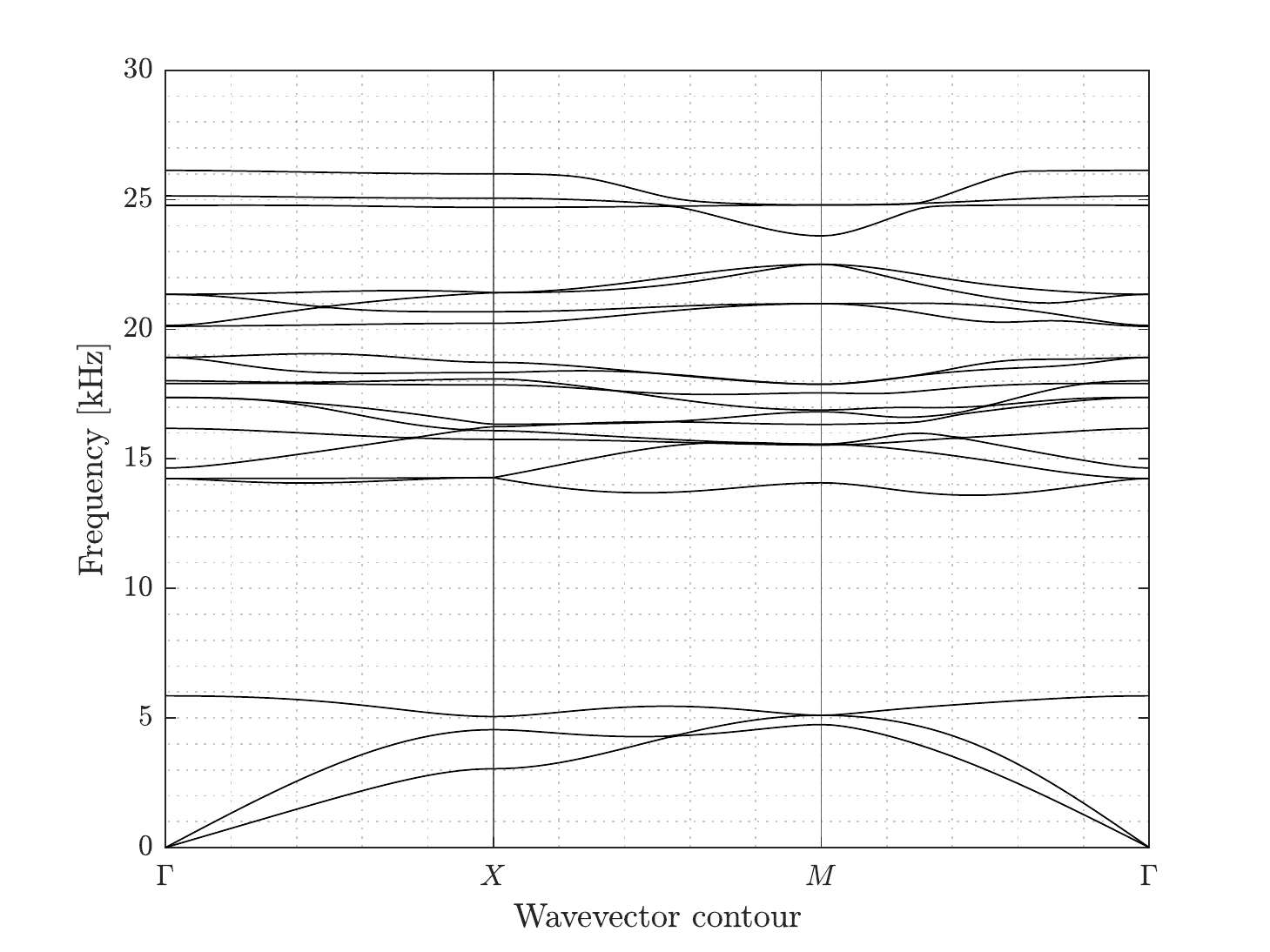}
    \caption{Left: Wavevector contour (also known as Irreducible Brillouin Zone contour). Right: Example of a dispersion relation evaluated along the wavevector contour.}
    \label{fig:wavevector}
\end{figure}

\subsection*{Commercial/Finite-medium Simulator}

Dispersion relation analysis is performed only on the domain of a single unit-cell, and assumes that the metamaterial has a unit cell that is infinitely repeating in each direction that Bloch-Floquet periodic boundary conditions are applied. However, real periodic metamaterials do not extend infinitely in any directions. While dispersion analysis is a critically useful method for converging on a unit-cell design with desirable properties, it is important to check that these properties are still present when the metamaterial is only finitely periodic.

To perform this validation, we perform frequency domain analysis on finitely periodic geometries using the commercial Finite Element software \textit{COMSOL Multiphysics} \cite{44} with the Structural Mechanics Module.

\subsection*{Material Properties}
The material is a 2-D metamaterial made by tiling a $10\times 10$ pixelated unit-cell. Each unit-cell is made of two constituent materials: one is soft and light, with elastic modulus and density of $E = 2 [GPa]$, $\rho = 1,000 [kg/m^3]$ respectively, and the other is stiff and heavy with $E = 200 [GPa]$, and $\rho = 8,000 [kg/m^3]$.

\section{Relationship of Proposed Methods to Other Machine Learning Methods}
\label{sec:related_work}
\textit{Comparison of shape-frequency features to CNN filters}: One might think that the sliding windows of the shape-frequency features are similar to convolution filters in CNNs, but they are not. Shape-frequency features count only  exact matches between each window and the soft constituent material, while a traditional convolutional layer would output a real number for the degree of match. In order to attempt to replicate the calculation of shape-frequency with a CNN, one would  concatenate a convolutional layer, a Rectified Linear Unit (ReLU) layer and a linear layer in a CNN \cite{krizhevsky2012imagenet}: the convolutional layer calculates the matching score, the ReLU layer sets a threshold to round the matching score to 0 or 1, and the linear layer averages the rounded score over all locations. However, if we train such a CNN, the weights cannot be integers and thereby the model cannot be as sparse and interpretable as our method using the shape-frequency features.

\textit{Comparison of \PrototypeModel{s} to prototype learning and decision sets:}
The \PrototypeModel{} technique is closely related to prototype-based machine learning models (e.g., \cite{bien2011prototype,kim2014bayesian}). The prototype-based models learn a set of prototypical cases (which can be whole observations or parts of  observations) from the training data. Given a test observation, the model makes a decision by comparing the test observation to the prototypical cases. For example, in classification, the model would output the label of the the prototypical case that most closely resembles the test observation. A \Prototype{} can be viewed as a part-based prototype because it is part of a training observation, and the decision of the model is made by matching the templates and the test observation. A difference between classical prototype-based learning and our approach is that classical objectives are classification losses while our method optimizes support and precision.

A \PrototypeModel{} can also be viewed as an instance of a decision set. A decision set (e.g., \cite{clark1991rule,frank1998generating,WangEtAl16,lakkaraju2016interpretable,WangEtAl2017}), is a logical model comprised of an unordered collection of rules, where each rule is a conjunction of conditions. In other words, the entire model is a disjunctive normal form (DNF), or ``OR of ANDs.'' A positive prediction is made if at least one of the rules is satisfied. One \Prototype{} in the set is essentially a conjunction of conditions that compares whether the pixel in the template is the same as the pixel in the test observation. That is, there is an ``OR'' over unit-cell templates and an ``AND'' over pixels in the template. As far as we know, decision sets have not previously been utilized for material discovery. 
\section{Visual Illustration of Sampling in Finer Resolution Design Space}

\begin{figure*}
    \centering
    \begin{subfigure}[t]{1\textwidth}
    \centering
    \includegraphics[width=5.0in]{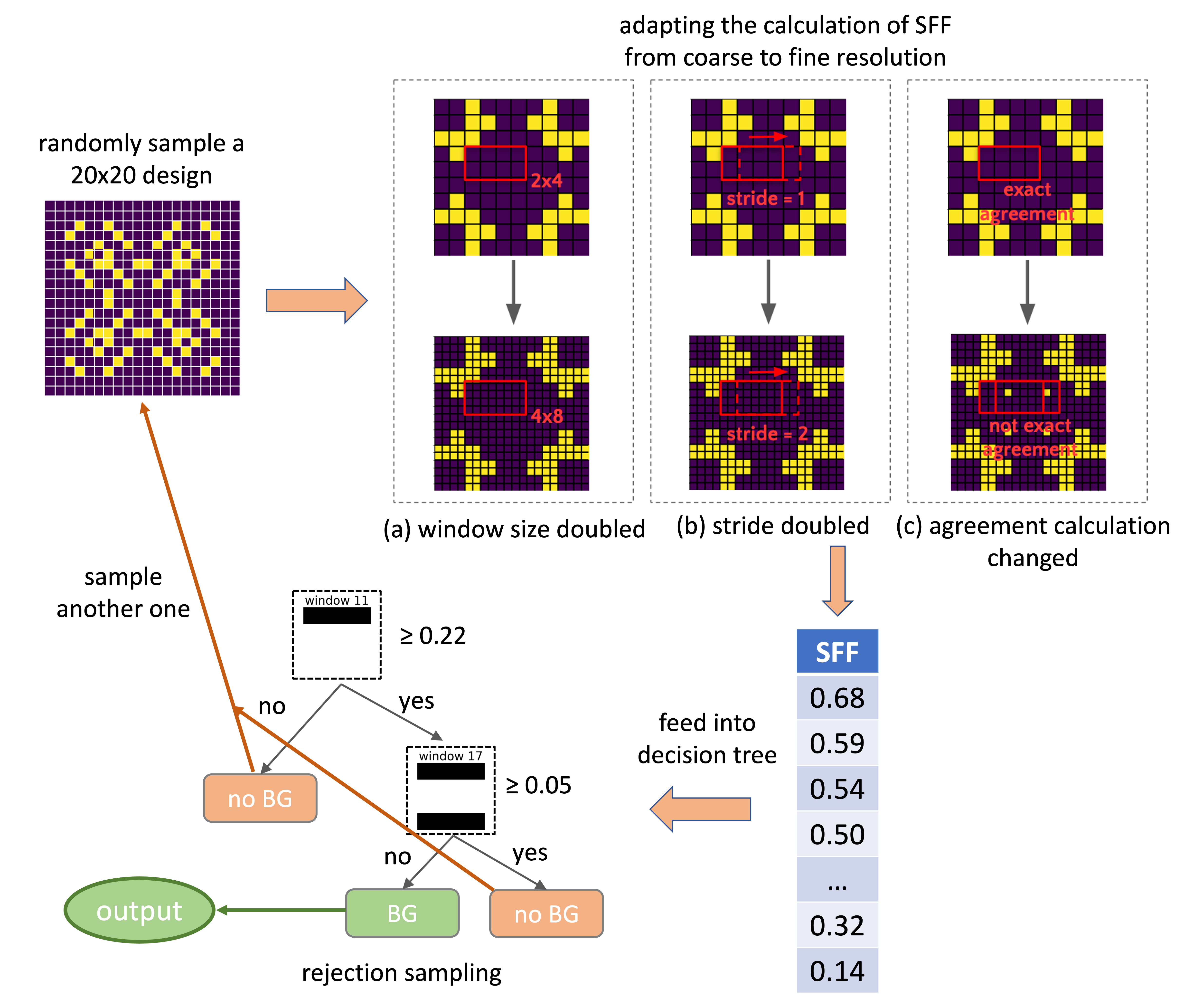}
    \caption{Sampling finer resolution design with SFF+GOSDT trained on coarse resolution data. Sampling consists of rejection sampling with GOSDT, and adapting the calculation of SFF from coarse to fine resolution (see Section 4.1.3).}
    \end{subfigure}
    \begin{subfigure}[t]{1\textwidth}
    \centering
    \includegraphics[width=5.0in]{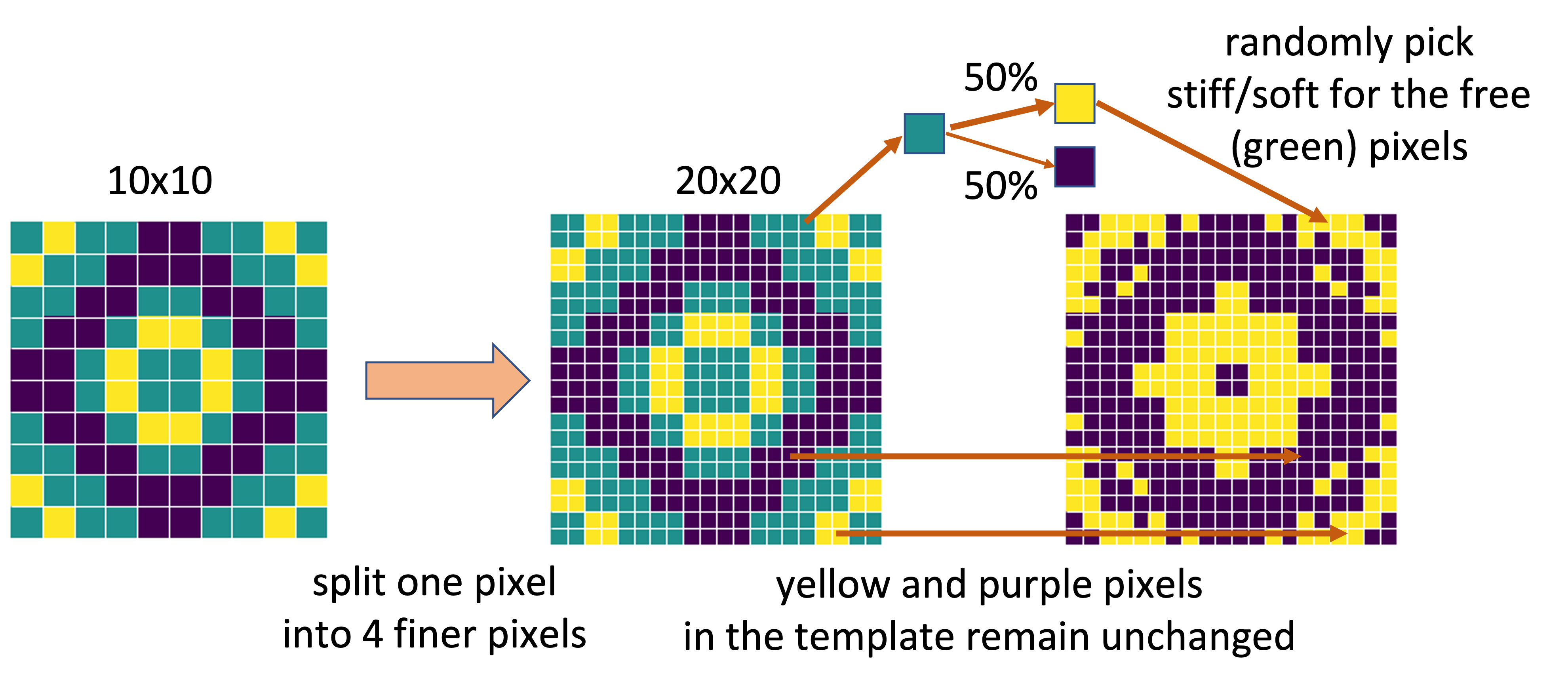}
    \caption{Sampling finer resolution design with \Prototype{} trained on coarse resolution data.}
    \end{subfigure}
    \caption{Visual illustration of how to sampling finer resolution design with proposed methods.}
    \label{fig:finer_scale}
\end{figure*}

Figure \ref{fig:finer_scale}(a) shows an example of how to sample a $20\times 20$ design with GOSDT+SFF trained on $10\times 10$ data. This is essentially a rejection sampling process, where a random $20\times 20$ design is selected, and accepted if the algorithm predicts existence of a band gap, and rejected if not. However, as the design space has changed from $10\times 10$ to $20\times 20$, the calculation of SFF has also changed, with details discussed in Section 4.1.3 in the main text.

Figure \ref{fig:finer_scale}(b) shows an example of how to sample a $20\times 20$ design with a $10\times 10$ \Prototype{}. While still following the templates learned from coarse resolution ($10\times 10$) data, most of these finer resolution designs can never be found in the coarse resolution ($10\times 10$) design space.

\section{More Experimental Results}

\subsection{Structure-to-property - Prediction Accuracy}
\begin{figure*}[h]
    \centering
    \begin{subfigure}[t]{0.3\textwidth}
    \includegraphics[scale=0.28]{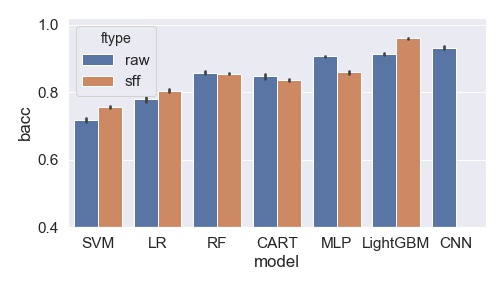}
    \caption{freq.$\in$\lbrack0, 10\rbrack\ kHz}
    \end{subfigure}
    ~
    \begin{subfigure}[t]{0.3\textwidth}
    \includegraphics[scale=0.28]{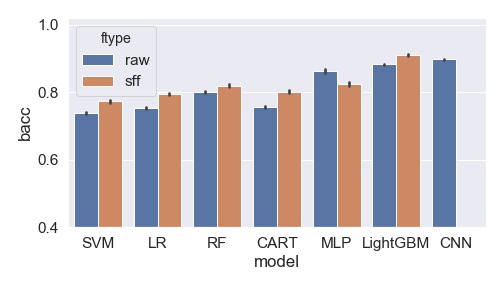}
    \caption{freq.$\in$\lbrack10, 20\rbrack\ kHz}
    \end{subfigure}
    ~
    \begin{subfigure}[t]{0.3\textwidth}
    \includegraphics[scale=0.28]{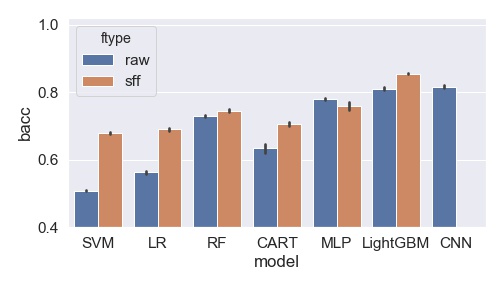}
    \caption{freq.$\in$\lbrack20, 30\rbrack\ kHz}
    \end{subfigure}
    \vskip\baselineskip
    \begin{subfigure}[t]{0.3\textwidth}
    \includegraphics[scale=0.28]{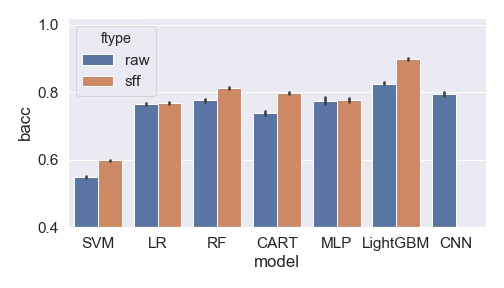}
    \caption{freq.$\in$\lbrack30, 40\rbrack\ kHz}
    \end{subfigure}
    ~
    \begin{subfigure}[t]{0.3\textwidth}
    \includegraphics[scale=0.28]{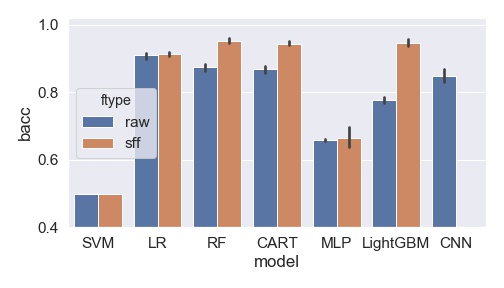}
    \caption{freq.$\in$\lbrack40, 50\rbrack\ kHz}
    \end{subfigure}
 \caption{Testing balanced accuracies (bacc) of different machine learning models. The label for each machine learning problem is in its subfigure caption, for instance, freq.$\in$\lbrack10, 20\rbrack\ kHz means predicting whether the material has a band gap within any frequency in the range \lbrack10, 20\rbrack\ kHz. Using the shape-frequency features (sff) generally improve performance over raw features. Note that it is not natural to design a convolution over the shape-frequency features since there is no notion of adjacency, so there is no CNN value for the shape-frequency features.}
\label{fig:baccs}
\end{figure*}

We now compare the testing balanced accuracy (bacc) of various machine learning methods, including linear models like support vector machines with linear kernels (SVMs) and logistic regression (LR); tree-based models like CART, random forest (RF), and boosted trees (LightGBM \citep{ke2017lightgbm}); as well as neural networks including the multi-layer perceptron (MLP) and convolutional neural networks (CNNs). LightGBM \citep{ke2017lightgbm} was used with 300 trees, and the MLP had 2 hidden layers, each containing 100 neurons. The CNN architecture is ResNet18 \citep{he2016deep} adapted to one-channel input, and default parameters were used for other methods. All other models were trained on shape-frequency features (SFF) and raw features respectively, while the CNN was trained on the entire 2-D design. We trained each model 5 times to average out randomness in the training process. Figure \ref{fig:baccs} shows the balanced accuracy results. Models trained on shape-frequency features generally perform much better than models trained on raw features, with the exception of MLP. Also, LightGBM trained on SFF outperforms CNN.

\subsection{Property-to-structure Sampling - Precision and Support}


\begin{table*}[h]
    \centering
    \begin{tabular}{|c|c|c|c|}
    \hline Raw+LightGBM & SFF+LightGBM & SFF+GOSDT (ours) & Unit-Cell Template Sets (ours) \\\hline
        $5.1\times10^{-6}$ s & $3.0\times10^{-4}$ s & $1.8\times10^{-4}$ s& \bm{$1.9\times10^{-7}$} \textbf{s} \\\hline
    \end{tabular}
    \caption{Average sampling time of a new $10\times10$ design.
    }
    \label{tab:sampletime_10x10}
\end{table*}

Table \ref{tab:sampletime_10x10} compares the average time per sample drawn for each method. \textbf{The \PrototypeModel{} has the fastest sampling speed} because the model explicitly represents all designs that match with it, while other methods rely on rejection sampling to solve the inverse problem. Models trained on shape-frequency features have a slower sampling rate than models trained on raw features, since calculation of shape-frequency features takes extra time. Also, due to their sparsity, GOSDT models have faster sampling than LightGBM. However, all these the sampling times work fine in practice.

\begin{table*}[h]
    \centering
    \begin{tabular}{|c|c|c|c|}
        \hline
         Frequency range & $s=5$ & $s=8$ & $s=10$\\\hline
         \lbrack0, 10\rbrack\ kHz & 98.53\%, 339& 98.54\%, 481& 97.73\%, 529\\  \hline
         \lbrack10, 20\rbrack\ kHz & 98.68\%, 758 & 98.60\%, 1142& 98.35\%, 1274\\\hline
         \lbrack20, 30\rbrack\ kHz & 94.08\%, 203 & 94.77\%, 287& 95.33\%, 343\\\hline
    \end{tabular}
    \caption{Testing precision and support of \PrototypeModel{s} with different sparsity constraints. $s$ is the number of templates. Numbers in the table cells are formatted as ``precision, support.'' The testing support here is calculated among 6554 testing samples (20\% of the entire dataset).}
    \label{tab:precision_support_s}
\end{table*}
Table \ref{tab:precision_support_s} compares the performance of \PrototypeModel{s} with different sparsity constraints ($s=5$, $s=8$, and $s=10$). The MIP corresponding to each model was run with a time limit of 60 minutes. The best solution at that expiration time was used if the MIP was not solved to optimality. The results suggest that there is no significant precision difference when using different numbers of templates in the model. The support, i.e., the number of designs covered by the template set, increases as the number of templates increase. In practice, a support of $s=5$ tends to be good enough.

\subsection{Transfer to Finer Resolution - Transfer Precision}

\begin{figure}[!htp]
    \centering
    \begin{subfigure}[t]{1\textwidth}
    \includegraphics[scale=.38]{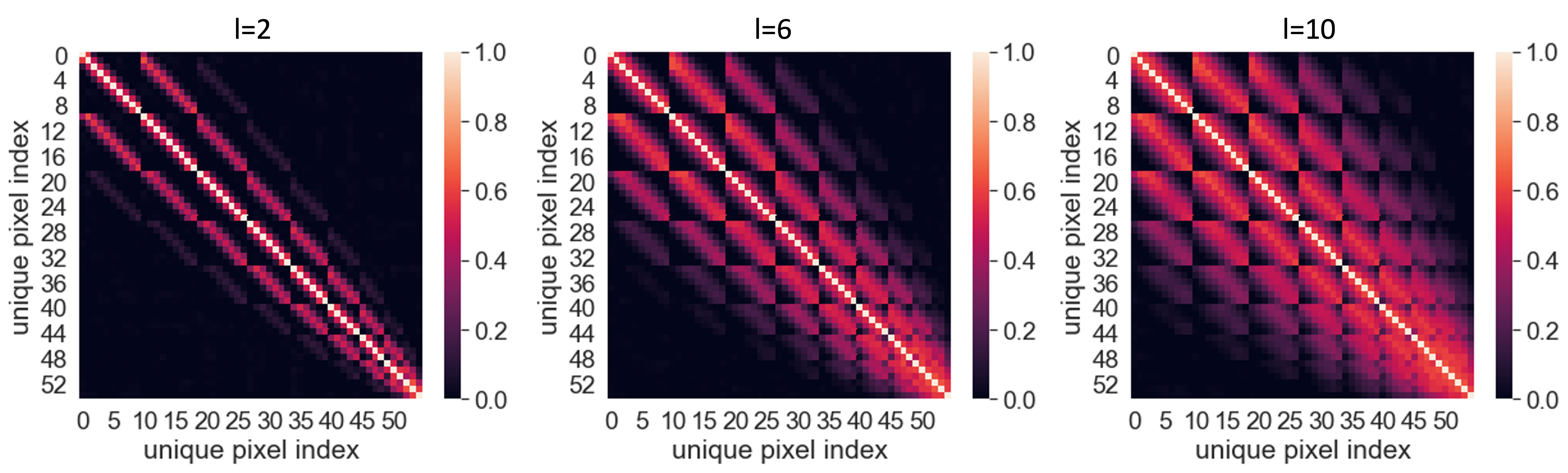}
    \caption{Matern 3/2 covariance matrices with different scale parameter ($l=2$, $l=6$, and $l=10$). The larger $l$ is, the more correlated the design pixels are.}
    \end{subfigure}
    \begin{subfigure}[t]{1\textwidth}
    \includegraphics[scale=.38]{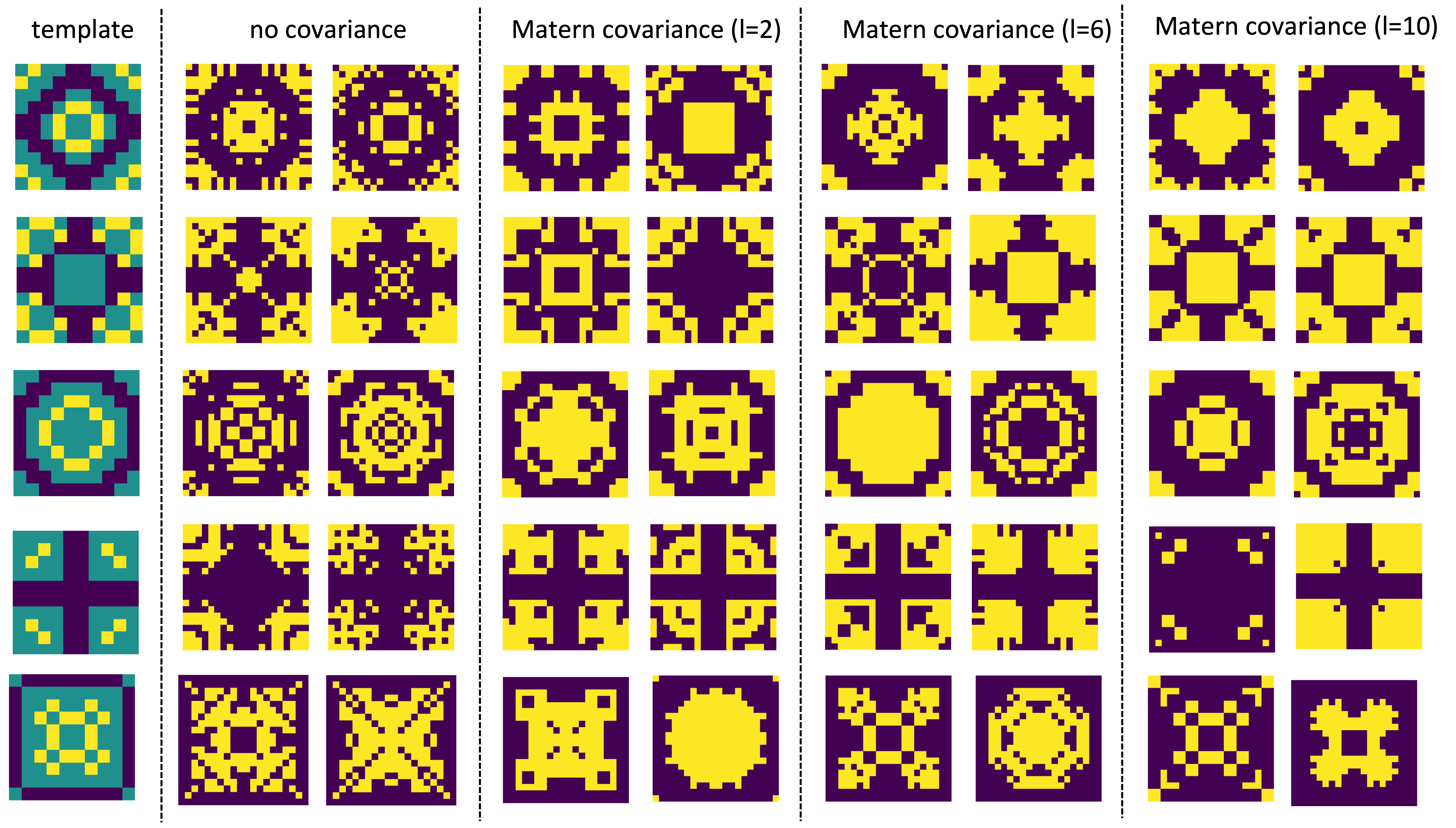}
    \caption{$20\times 20$ designs sampled from \PrototypeModel{} for band gaps in \lbrack0, 1000\rbrack{} using different sampling methods. From left to right are the template, pixels sampled with no covariance (independently), Matern 3/2 covariance ($l=6$), Matern 3/2 covariance ($l=2$), and Matern 3/2 covariance ($l=10$)}
    \end{subfigure}
    \caption{Sampling with correlation. (a) covariance matrices; (b) sampled finer resolution ($20\times 20$) designs} \label{fig:correlation}
\end{figure}

In order to test the flexibility of \PrototypeModel{s} in sampling finer scale designs, in addition to sampling the pixels independently in the free region of a \Prototype{}, as we did in the main paper, we also try to sample pixels with correlation between them. Specifically, we use a Matern $\frac{3}{2}$ kernel, i.e.,
\begin{equation}
k(\mathbf{x},\mathbf{x}') = (1+\sqrt{3}\cdot\frac{d(\mathbf{x},\mathbf{x}')}{l})\cdot e^{-\sqrt{3}\cdot\frac{d(\mathbf{x},\mathbf{x}')}{l}}
\end{equation}
to calculate the covariance between two pixels $\mathbf{x}$ and $\mathbf{x'}$; $d$ is the distance function, which is set to be the distance between the two pixels in the 2D design space; $l$ is the scale parameter, the larger $l$ is, the more the correlation spreads throughout the larger design region of the unit cell. Note that since the design pixels are binary, we cannot directly sample them with a given covariance matrix. Instead, we first rescale the covariance matrix elements by a transformation function $g(x) = \text{sin}(\frac{x\pi}{2})$, then sample from a multivariate Gaussian distribution, and take the sign of the Gaussian variables to get the binary variables/pixels. Figure \ref{fig:correlation}(a) shows the covariance matrices with different scale parameters, and Figure \ref{fig:correlation}(b) shows the designs whose free pixels (green region of templates) are sampled with (i) no covariance, and (ii) the Matern 3/2 covariance matrices with $l=2$, $l=6$, and $l=10$. Generally, the designs sampled with Matern covariance matrices are more regularized than independent sampling --- independent designs look like coarse templates with fine white noise, while the Matern covariance designs are more similar to designs generated by humans. Also, larger values of the length scale parameter $l$ would create larger length scale features (regions with larger areas of purple or yellow).

\begin{table}
    \centering
    \begin{tabular}{|c|c|c|c|c|}
        \hline Frequency range & no covariance  & Matern 3/2 ($l=2$) & Matern 3/2 ($l=6$) & Matern 3/2 ($l=10$) \\\hline
         \lbrack0, 10\rbrack\ kHz & 100.0\% & 98.0\%& 96.5\% & 96.5\% \\\hline
         \lbrack10, 20\rbrack\ kHz & 98.5\% & 98.5\% & 96.5\% & 98.0\% \\\hline
         \lbrack20, 30\rbrack\ kHz & 91.0\% & 92.5\% & 95\%  & 94\%\\\hline
    \end{tabular}
    \caption{Transfer precision from coarse resolution ($10\times 10$) to finer resolutions with different sampling methods ($20\times 20$) in the free regions of the unit cell templates.}
    \label{tab:correlation}
\end{table}
Table \ref{tab:correlation} shows the precision of these $20\times 20$ designs. Interesting phenomena about wave physics can be founding when comparing the precisions in different frequency ranges and feature sizes of the generated designs. In the lowest frequency range, i.e., [0, 10] kHz, the transfer precision is the lowest at $l=10$, when the generated designs have largest feature length scales. Here, length scale of feature means the average size of regions with the same color. In the highest frequency range, designs with the smallest feature length scale (independent fine pixels) produce the lowest precision.  In the middle frequency range [10, 20] kHz, precision is the lowest at $l=6$, which corresponds to medium feature length scales. This agrees with intuition about band gap wave physics --- band gaps in higher frequency ranges are more closely associated with finer-scale features, and thus would naturally be more highly affected by finer-scale features generated in the green regions of unit-cell templates. Surprisingly, the unit cell templates still perform well (lowest precision $>$ 90\%) even considering the interplay between wavelength and length scale of feature. These results indicate that our unit cell templates are very robust and flexible for material design. Given a template, materials scientists can design features almost freely in the green region to satisfy  practical needs (e.g., connectivity constraints) while still maintaining the desired band gap. 

\subsection{Finite Tiling COMSOL Test}
We use a commercial finite-element solver \citep[COMSOL,][]{multiphysics1998introduction} to simulate the dispersion relations of finitely tiled materials.
COMSOL is a standard, industrial finite element software tool. We set the label to 1 when the size of the band gap in the target frequency range was greater than 1 kHz; these large band gaps are useful in practice.

\begin{figure}[!htp]
    \centering
    \begin{subfigure}[t]{0.9\textwidth}
    \includegraphics[scale=0.5]{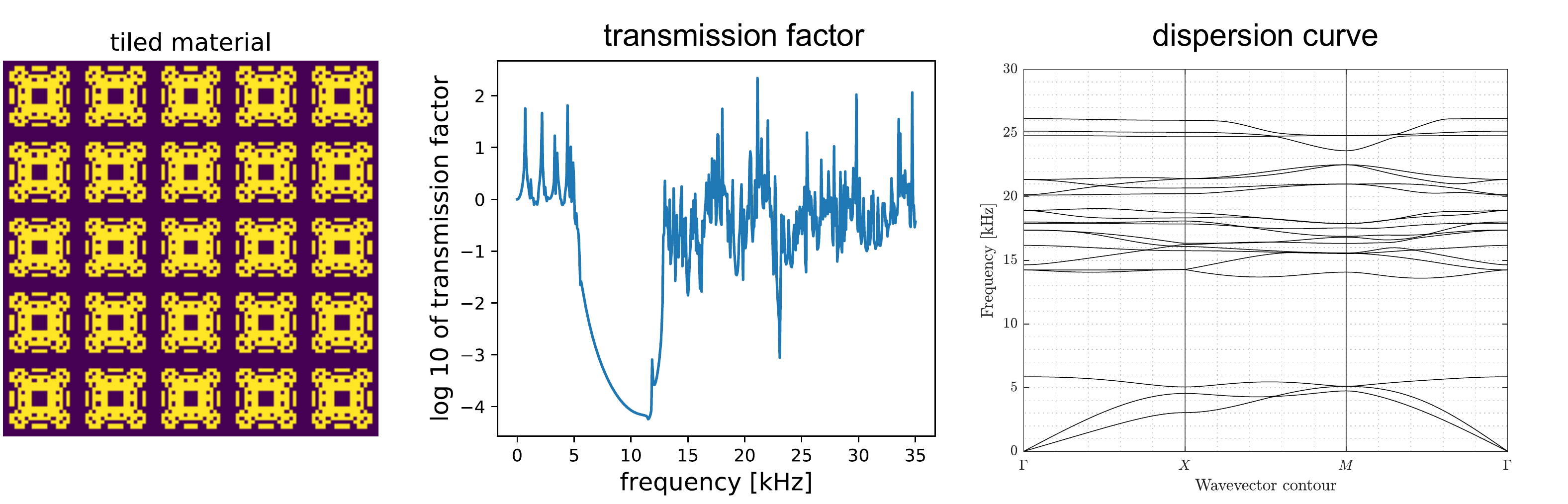}
    \caption{Frequency range: \lbrack0,10\rbrack\ [kHz] }
    \end{subfigure}
    ~
    \begin{subfigure}[t]{0.9\textwidth}
    \includegraphics[scale=0.5]{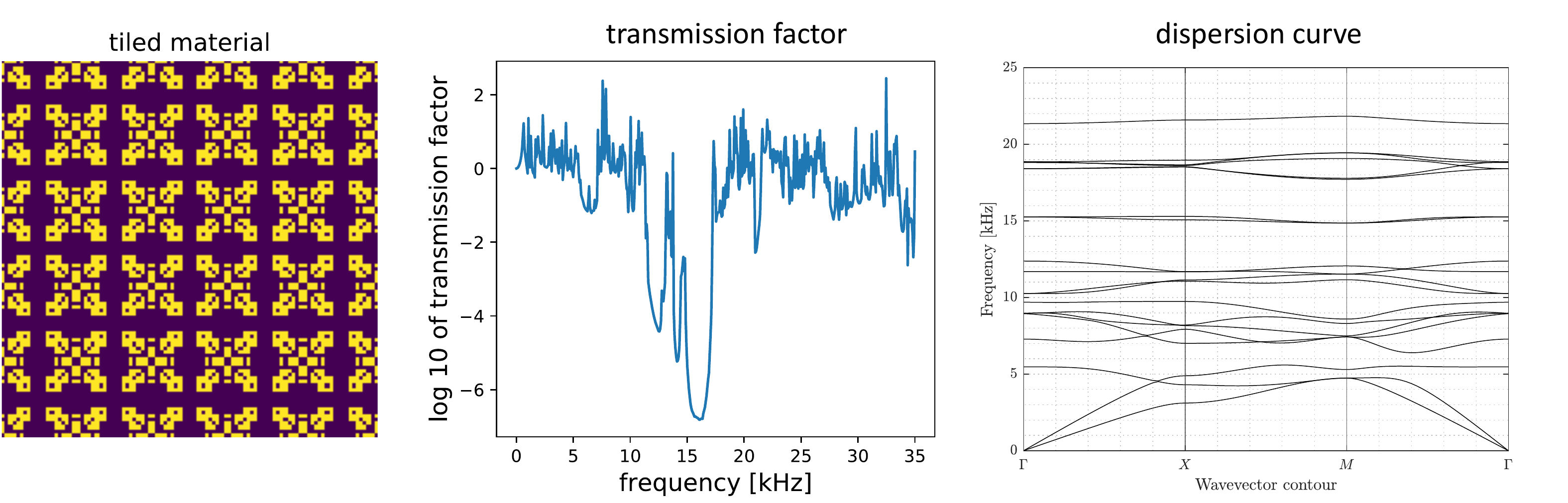}
    \caption{Frequency range: \lbrack10,20\rbrack\ [kHz]}
    \end{subfigure}
    \vskip\baselineskip
    \begin{subfigure}[t]{0.9\textwidth}
    \includegraphics[scale=0.5]{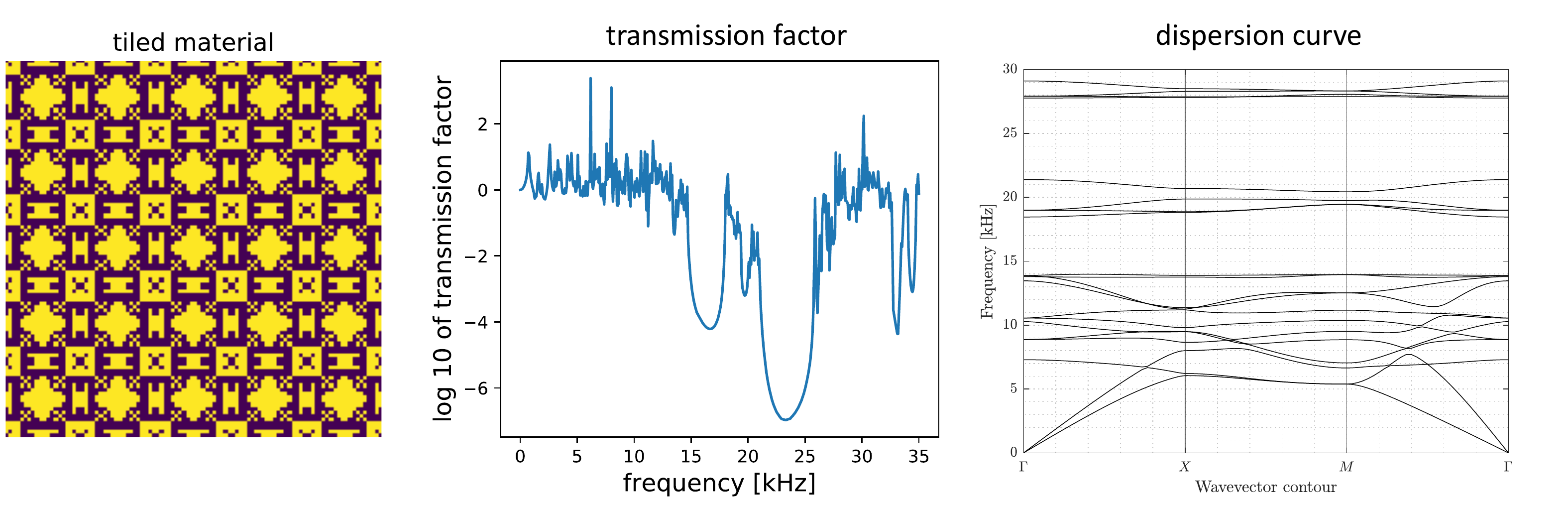}
    \caption{Frequency range: \lbrack20,30\rbrack\ [kHz]}
    \end{subfigure}
    \caption{Finitely tiled materials and their corresponding transmission factor. \textit{Left}: Tiled material (tiled $5\times 5$ times) made by unit-cell designs selected by the proposed method with respect to three different frequency ranges. \textit{Middle}: the transmission factor is the ratio between the magnitude of the response on the right side of the material to the input signal on the left side of the material. \textit{Right}: the dispersion curves of the tiled material.}
\label{fig:tiled_test}
\end{figure}

Figure \ref{fig:tiled_test} shows three examples of finitely tiled materials made by $20\times 20$ unit-cells found by the \PrototypeModel{s}, along with the wave transmission factors simulated by COMSOL, and their dispersion curves. The transmission factor is the ratio between the magnitude of the response on the right side of the material to the input signal on the left side of the material. Sharp drops in the transmission factor happen in the target frequency ranges in all three examples, indicating the existence of a band gap, which matches with the band gaps shown between the dispersion curves. In addition to the three examples in Figure \ref{fig:tiled_test}, we tested the same finite tiling for 100 unit-cells in each frequency range, and all of them have the target band gap property. Therefore, our method is robust under finite tiling.

\section{Extension to Non-square Pixels and Unit-cells}
\begin{figure*}[th]
    \centering
    \includegraphics[width=6in]{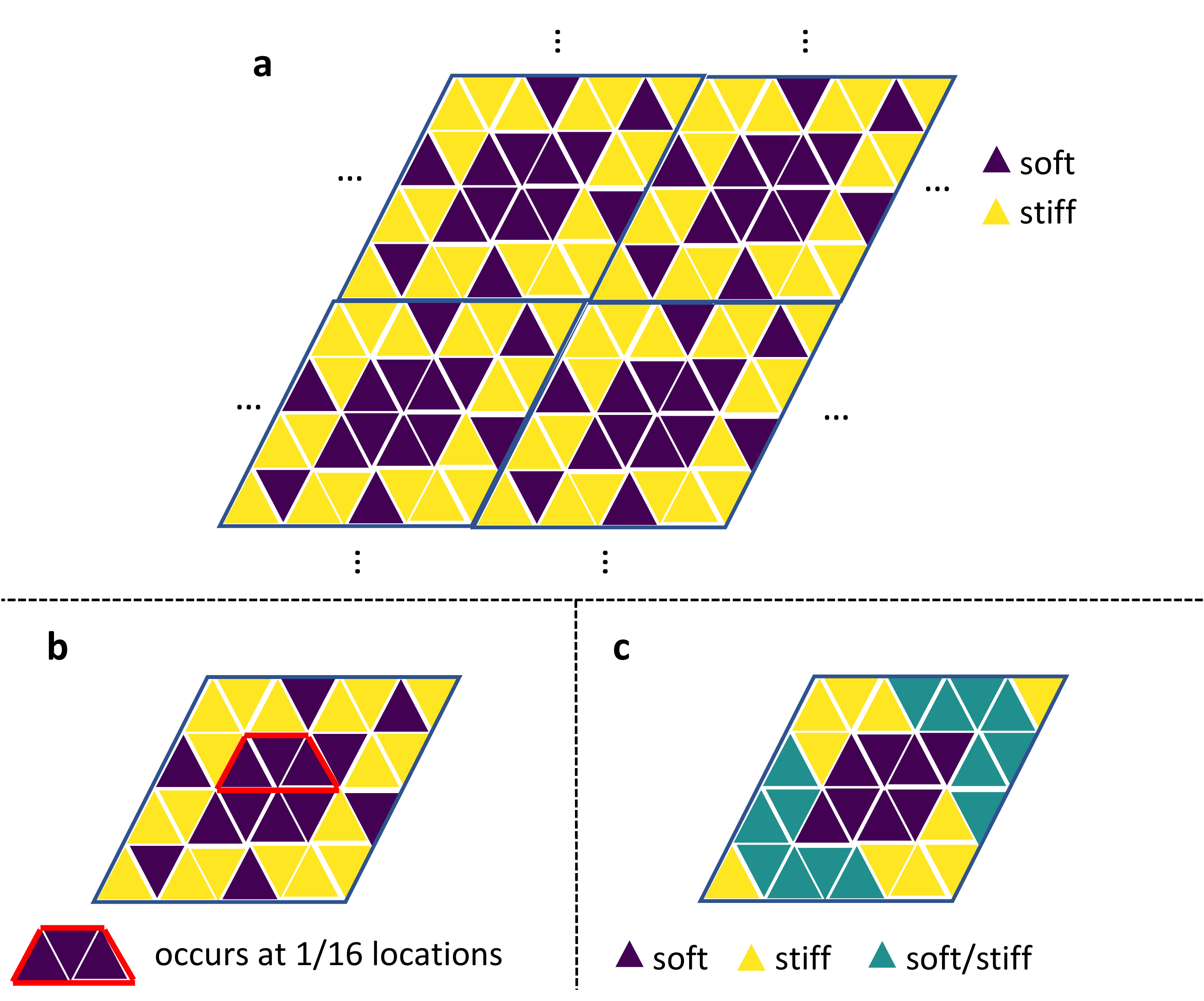}
    \caption{Nonsquare unit-cell and adaptation of proposed methods. \textbf{a} An example of metamaterial unit-cell design with triangle pixel and diamond unit-cell \textbf{b} An example of shape-frequency feature defined on this nonsquare design space; \textbf{c} A unit-cell template defined on this nonsquare design space that matches with the example unit-cell in \textbf{a}.}
    \label{fig:nonsquare}
\end{figure*}

In the problem setting of our experiments, both the pixels and unit-cells are square. This is mainly due to the simplicity of simulation in square settings. However, the proposed methods are general enough to be adapted to nonsquare pixels and unit-cells. Our methods try to calculate or identify the existence of certain global or local patterns, but do not specify that their shape should be comprised of square pixels. As an example, Figure \ref{fig:nonsquare} shows how shape-frequency features and unit-cell templates can be adapted to nonsquare pixels (triangle pixel) and unit-cells (diamond unit-cell).

\vfill

\end{document}